\documentclass[sigconf]{acmart}
\usepackage{paralist}
\usepackage{multirow}
\usepackage{graphicx}
\usepackage{subcaption}
\usepackage{booktabs}
\usepackage{amsthm}
\usepackage{makecell} 
\usepackage{amsmath}
\usepackage{algorithm}
\usepackage{algpseudocode}
\usepackage{geometry}
\newtheorem*{theorem-nonumber}{Theorem}
\usepackage{balance}
\AtBeginDocument{%
  }

\copyrightyear{2026}
\acmYear{2026}
\setcopyright{cc}
\setcctype{by}
\acmConference[KDD 2026] {Proceedings of the 32nd ACM SIGKDD Conference on Knowledge Discovery and Data Mining V.1}{August 9--13, 2026}{Jeju Island, Republic of Korea.}
\acmBooktitle{Proceedings of the 32nd ACM SIGKDD Conference on Knowledge Discovery and Data Mining V.1 (KDD 2026), August 9--13, 2026, Jeju Island, Republic of Korea}
\acmISBN{979-8-4007-2258-5/2026/08}
\acmDOI{10.1145/3770854.3780217}

\begin{document}

{
\title{Scalable Graph Condensation with Evolving Capabilities}

\author{Shengbo Gong}
\authornote{Both authors contributed equally to this research.}
\email{shengbo.gong@emory.edu}
\affiliation{%
  \institution{Emory University}
  \city{Atlanta}
  \state{Georgia}
  \country{USA}
}

\author{Mohammad Hashemi}
\email{mohammad.hashemi@emory.edu}
\authornotemark[1]
\affiliation{%
  \institution{Emory University}
  \city{Atlanta}
  \state{Georgia}
  \country{USA}
}

\author{Juntong Ni}
\email{juntong.ni@emory.edu}
\affiliation{%
  \institution{Emory University}
  \city{Atlanta}
  \state{Georgia}
  \country{USA}
}
\author{Carl Yang}
\email{j.carlyang@emory.edu}
\affiliation{%
  \institution{Emory University}
  \city{Atlanta}
  \state{Georgia}
  \country{USA}
}
\author{Wei Jin}
\email{wei.jin@emory.edu}
\affiliation{%
  \institution{Emory University}
  \city{Atlanta}
  \state{Georgia}
  \country{USA}
}

\begin{abstract}
The rapid growth of graph data creates significant scalability challenges as most graph algorithms scale quadratically with size. To mitigate these issues, Graph Condensation (GC) methods have been proposed to learn a small graph from a larger one, accelerating downstream tasks. However, existing approaches critically assume a static training set, which conflicts with the inherently dynamic and evolving nature of real-world graph data. This work introduces a novel framework for continual graph condensation, enabling efficient updates to the distilled graph that handle data streams without requiring costly retraining.
This limitation leads to inefficiencies when condensing growing training sets.
In this paper, we introduce GECC (\underline{G}raph \underline{E}volving \underline{C}lustering \underline{C}ondensation), a scalable graph condensation method designed to handle large-scale and evolving graph data. GECC employs a traceable and efficient approach by performing class-wise clustering on aggregated features. Furthermore, it can inherit previous condensation results as clustering centroids when the condensed graph expands, thereby attaining an evolving capability. This methodology is supported by robust theoretical foundations and demonstrates superior empirical performance. Comprehensive experiments including real world scenario show that GECC achieves better performance than most state-of-the-art graph condensation methods while delivering an around 1000$\times$ speedup on large datasets.
\end{abstract}







\maketitle

\section{Introduction} \label{sec:intro}


Graph-structured data has become indispensable in various domains, including social networks~\citep{fan2019graph}, epidemiology~\citep{liu2024review}, and recommendation systems \citep{wu2020comprehensive}. The ability of graphs to represent complex relationships and dependencies has propelled their adoption in machine learning, especially with the advent of graph neural networks (GNNs) \citep{bronstein2017geometric}. However, the exponential growth of real-world graph datasets presents significant computational challenges, as the cost of training GNNs increases with the number of nodes and edges~\citep{hu2021ogb, complexity}. To address this, graph condensation (GC) techniques~\citep{jin2021graph, jin2022condensing, gcsntk, geom, sfgc, xiao2024simple} have been developed, which aim to produce significantly smaller yet information-rich graphs that accelerate GNN training.  For example,  GCond~\cite{jin2021graph}  condenses the Flickr dataset to 0.1\% of its original size while preserving 99.8\% of the original accuracy on node classification. These condensed graphs not only save significant storage space and transmission bandwidth, but also enhance the efficiency of retraining neural networks in many critical applications such as continual learning~\cite{liu2023cat}.

\begin{table}[t!]
\centering
\caption{Condensation time on evolving graphs, reported as a multiple of GNN training time (21.37 s) on uncondensed graphs (\textit{Reddit}, 232,965 nodes) in the final time step. The GNN training time on condensed graphs is 1.1 s.}
\label{tab:preliminary}
\resizebox{0.9\linewidth}{!}{
\begin{tabular}{lccccc}
\toprule
Method         & \textit{T$_1$}               & \textit{T$_2$}               & \textit{T$_3$}               & \textit{T$_4$}               & \textit{T$_5$}               \\ \midrule
\textbf{GCondX}&75×& 112×& 149×& 104×& 185×\\
\textbf{GCondX-Init}    & 75×& 89×& 107×& 120×& 134×\\
\textbf{GCondX-Grow}  & 75×& 94×& 112×& 101×& 119×\\
\bottomrule

\end{tabular}}
\end{table}

Despite the promise, existing GC methods face three significant limitations that limit their practical utility in real-world scenarios.

\textbf{(a) High Computational Overhead:} Most GC methods entail a computationally intensive condensation process, marked by repeated gradient calculations and updates.  This process closely resembles full GNN training, creating a paradox as it contradicts the primary objective of GC methods to enhance the efficiency of GNN training.
For instance, gradient matching techniques~\cite{jin2021graph,jin2022condensing} require consistent alignment of GNN gradients across successive iterations. Similarly, trajectory matching methods~\cite{sfgc,geom} necessitate the alignment of GNN parameters at various stages during the training process. These methods inherently demand significant GNN training and gradient processing, thus slowing down the procedure as the size of graph datasets expands.

\textbf{(b) Challenges with Evolving Graphs:} Existing GC methods are designed for static graphs and overlook the evolving nature of real-world graphs. In practice, graphs evolve over time, with nodes, edges, or attributes being added, removed, or modified. 

Existing graph condensation (GC) methods require re-running the entire condensation process from scratch whenever the training set changes, since modifications affect every synthetic node. Moreover, while synthetic graphs are expected to grow proportionally with the original graphs, an effective strategy for adapting to this growth is currently lacking.
In Table~\ref{tab:preliminary}, we report the relative condensation times for evolving graphs on the \textit{Reddit} dataset, using GCondX~\cite{jin2021graph} as a representative example. We compare three variants: (i) \textbf{GCondX}: Performs condensation from scratch at each time step. (ii) \textbf{GCondX-Init}: Retains synthetic nodes from the previous time step as initialization, updating all nodes during the condensation process. (iii) \textbf{GCondX-Grow}: Freezes the synthetic nodes from the previous time step and concatenate new synthetic nodes to grow the condensed graph.
The table indicates that the condensation process requires over 100$\times$ more time than the GNN training on the uncondensed graph. This substantial computational overhead suggests that current GC methods may not be practical for evolving graphs. Even though simple evolving strategies can reduce marginal time, the cost remains significantly higher than that of GNN training, underscoring the urgent need for a novel GC method that can efficiently handle graph evolution.

\textbf{(c) Lack of Traceability:} Many GC methods synthesize condensed graphs without explicitly establishing a connection between the nodes in the condensed graph and those in the original graph. This lack of explicit connections obscures the contribution of each original node to each condensed node, diminishing the traceability of the condensation process. Although these condensed nodes might be informative to GNN models, discerning their real-world semantic meanings can be challenging to humans. This poor traceability can significantly limit the application of GC in sectors where clarity, comprehensive data explanations, and transparent decision-making are crucial. Also, with traceability, users can manipulate the condensed graph alongside the original one—for example, easily filtering low-quality data when the corresponding original nodes exhibit poor quality.

To tackle the three critical challenges all together, we introduce a novel framework called \underline{G}raph \underline{E}volving \underline{C}lustering \underline{C}ondensation (GECC). We begin by refining our objectives for graph condensation and establishing theoretical upper bounds that clarify the connection between graph condensation and clustering. This connection enables us to perform condensation efficiently without the costly gradient calculations needed for traditional GNN training. Specifically, we cluster propagated node features into partitions, where centroids serve as the condensed node features at each time step. Additionally, this method adapts well to evolving, expanding graphs through incremental clustering, using centroids from previous time steps as initializations for subsequent steps. Notably, GECC enhances traceability by establishing a clear correspondence between condensed and original nodes, offering deeper insights into how condensed nodes encapsulate information from the actual graph. Our contributions are outlined as follows:

\begin{compactenum}[\textbullet]
    \item We provide a novel theoretical understanding of objectives in graph condensation from both the training and test perspectives. Our analysis reveals the connection between graph condensation and clustering, demonstrating that condensation can be effectively addressed by performing clustering with balanced clusters.
 
    \item Leveraging our theoretical insights, we introduce the first framework designed to significantly accelerate condensation and adapt to the evolving nature of real-world graphs, while providing traceability. This framework methodically partitions node representations to produce cluster centroids at each time step, which serve as the condensed node features.
    
    \item Comprehensive experiments demonstrate that GECC achieves state-of-the-art (SOTA) accuracy and efficiency in GC. For example, GECC can condense the evolving \textit{Reddit} dataset more than 1000 times faster than GCond, while outperforming other existing methods in terms of accuracy.
\end{compactenum}

\section{Related Work} 

\subsection{Model-based Graph Condensation}
Model-based GC methods, which require a GNN training phase on the original graph, were the first approaches proposed for GC. In these methods, a smaller graph is synthesized to effectively represent the original for training GNNs \citep{hashemi2024comprehensive}. For example, GCond \citep{jin2021graph}, DosCond \citep{jin2022condensing}, and SGDD \citep{yang2023does} minimize a gradient matching loss that aligns the gradients of the training losses w.r.t. the GNN parameters computed on both the original and condensed graphs. Alternatively, SFGC \citep{sfgc} and GEOM \citep{geom} condense a graph by aligning the parameter trajectories of the original graph, thereby eliminating the need for explicit edge generation and reducing complexity. Although these approaches are effective, they require significant computational resources due to the multiple full GNN training runs on the original graph. To mitigate this cost, GCDM \citep{liu2022graph} and SimGC~\cite{xiao2024simple} adopt a distribution matching strategy by aligning sampled original node features with synthetic nodes, thus reducing the discrepancy between gradient matrices and the final output. Other methods such as GCSNTK~\cite{gcsntk} use GNTK~\cite{du2019gntk} to bypass the gradient update in the synthetic graph; however, such approaches currently show suboptimal performance according to recent benchmarks \citep{gong2024gc4nc,sun2024gc}. OpenGC~\cite{opengc} enhances graph condensation for evolving graphs by simulating structure-aware distribution shifts to extract temporal invariant patterns, thereby improving the generalization of GNNs in real-world scenarios. 
Overall, these model-based GC methods share common challenges including high computational overhead, scalability, and traceability issues. 
The recently proposed MCond~\cite{gao2024graph} explicitly learns a one-to-many node mapping from original nodes to synthetic nodes, thereby facilitating inductive representation learning and enhancing traceability; however, it still operates within the framework of gradient matching and remains highly complex.

\subsection{Model-Agnostic Graph Condensation}
While model-based approaches have advanced GC, they still struggle to scale to large-scale graph datasets (e.g., graphs with over one million nodes such as Ogbn-products~\citep{hu2021ogb}). To further reduce condensation time on such large-scale graphs, recent efforts have focused on \textbf{directly matching the representations} between the original and synthetic graphs, bypassing the need for a surrogate model, i.e., the GNN model used during the condensation. This strategy offers benefits in efficiency and generalizability to diverse downstream models. GCPA \citep{li2025a} first extracts structural and semantic information from the original graph, randomly samples the representations, and finally refines them using contrastive learning. Likewise, CGC \citep{gao2024rethinking} reduces the representation matching problem to a class partitioning task. In this approach, nodes within the same partition are merged, and a learning module is employed to weight different samples before they are aggregated into a synthetic node. \textbf{However}, despite its effectiveness, CGC's theoretical analysis does not establish a relationship between class partitioning and the GC objective, thereby questioning the necessity of this extra step. \textbf{Moreover}, these model-agnostic methods still require a \textit{learning module}, which introduces challenges in optimization and hyperparameter tuning. \textbf{Finally}, none of these methods fully address the evolving nature of large-scale graphs, which typically necessitates a complete re-condensation whenever the graph changes. \textit{To our best knowledge, GECC is the first model-agnostic and training-free method with linear complexity.}

\section{Preliminaries and Notations} 

 Given a node set $\mathcal{V}$ and an edge set $\mathcal{E}$, a graph is denoted as $G=({\mathcal{V},\mathcal{E}})$. In the case of attributed graphs, where nodes are associated with features, the graph can be represented as $G=({\bf{X}, \bf{A}})$, where ${\bf X}=[{\bf x}_1,{\bf x}_2,...,{\bf x}_N]$ denotes the node attributes, and ${\bf A}$ denotes the adjacency matrix. The graph Laplacian matrix is $\bf L=\bf D-\bf A$, where $\bf D$ is a diagonal degree matrix with ${\bf D}_{ii}=\sum_j {\bf A}_{ij}$. Let $N=|\mathcal{V}|$ and $E=|\mathcal{E}|$ represent the number of nodes and edges, respectively.

\subsection{Graph Condensation.}
GC aims to create a smaller synthetic graph \( G' = (\mathbf{X}', \mathbf{A}') \), where \( \mathbf{X}' \in \mathbb{R}^{N' \times d} \), \( \mathbf{A}' \in \{0, 1\}^{N' \times N'} \), and \( N' \ll N \), from the original large graph \( G = (\mathbf{X}, \mathbf{A}) \). The objective is to ensure that GNNs trained on \( G' \) achieve performance comparable to those trained on \( G \), thereby significantly accelerating GNN training \citep{jin2021graph}. 
The large-scale graph \( G = (\mathbf{X}, \mathbf{A}) \) serves as the original graph of our GC framework. Each node is annotated with one of \( c \) classes, encoded as numeric labels \( \mathbf{y} \in \{1, \dots, c\}^{N} \) and one-hot labels \( \mathbf{Y} \in \mathbb{R}^{N \times c} \). Any GC method focuses on finding an update function $\mathcal{F}$ that generates a new condensed graph $G' = \mathcal{F}(G)$, preserving the key structural and feature information required for downstream tasks. 

\subsection{Evolving Graph Condensation.} 
\textbf{Motivation}: Training GNNs on real-world graphs is challenging because these graphs constantly evolve, leading to significant computational costs. Current GC techniques, which create smaller training graphs, are designed for static snapshots. Consequently, adapting to graph evolution requires re-running the entire expensive condensation process at each step. This wasteful approach motivates our work on Evolving Graph Condensation, which aims to efficiently update a condensed graph incrementally rather than rebuilding it from scratch.
In the evolving graph scenario, we consider a sequential stream of graph batches \(\{B_1, B_2, \dots, B_m\}\), where \(m\) represents the total number of time steps. Each graph batch \(B_i = (\mathbf{X}_i, \mathbf{A}_i)\) contains newly added nodes along with their associated edges in \textit{inductive} graphs, while in \textit{transductive} graphs, it contains newly labeled nodes but retains the entire graph structure~\cite{su2023robustincremental}. 

\noindent\textbf{Problem Formulation}: In the evolving graph scenario, we consider a sequential stream of graph batches $\{B_1, B_2, \dots, B_m\}$, where $m$ is the total number of time steps. Each graph batch $B_i = (\mathbf{X}_i, \mathbf{A}_i)$ introduces new information. In inductive graphs, a batch contains newly added nodes and their edges. In transductive graphs, it typically contains newly revealed labels for existing nodes while the overall structure is retained~\cite{su2023robustincremental}. These batches are progressively integrated, constructing a series of snapshot graphs $\{G_{1}, G_{2}, \dots, G_{m}\}$. At any time step $t$, the snapshot graph $G_{t} = (\mathbf{X}_{t}, \mathbf{A}_{t})$ encompasses all information up to that point, where $G_{t} = \bigcup_{i=1}^t B_i$. Given the snapshot graph $G_t$ at the current time step $t$ and the condensed graph $G'_{t-1}$ from the previous step, the goal of evolving graph condensation is to find an update function $\mathcal{F}$ that generates a new condensed graph $G'_t = \mathcal{F}(G'_{t-1}, B_t)$. The objective is for $G'_t$ to be significantly smaller than $G_t$ while preserving essential structural and feature information, such that any GNN trained on $G'_t$ achieves performance on downstream tasks that is comparable to a GNN trained on the full, original graph $G_t$.
At each time step $t$, instead of repeating the GC process from scratch on $G_t$, our method aims to efficiently update the condensed graph. It effectively inherits knowledge from the previously condensed graph $G'_{t-1}$ and integrates the new information from batch $B_t$ to produce $G'_t$. This updated graph is then used to train GNNs, which are deployed to classify nodes in the full graph $G_t$. This incremental approach ensures that the condensed representation stays current with the evolving graph while minimizing redundant computation.


\section{Methodology}

In this section, we outline our condensation objectives for preserving training data information and generalizing to test data in static graphs. We employ theoretical analysis to simplify these objectives and introduce a scalable condensation method that does not require training GNN models, while offering traceability by establishing clear correspondence between the original and condensed nodes. Building on this scalable approach, we further propose an efficient adaptation to the evolving scenarios.

\subsection{A Deep Dive into Condensation Objectives}
Intuitively, the condensation process should preserve sufficient information from training data to maintain GNN training performance while ensuring model generalization to test data. Thus, we divide our discussion into two stages:  \textit{training} and \textit{test}.  To simplify our analysis, we adopt the {Simplified Graph Convolution} (SGC) model as the GNN~\cite{wu2019simplifying}, due to its simpler design, similar filtering behavior to GCN~\cite{kipf2016semi}, and its frequent use as a backbone and evaluation model in numerous graph condensation works~\cite{jin2021graph, jin2022condensing,gong2024gc4nc, xiao2024simple}.

\textbf{Training Stage Objectives.} 
During the training stage, GC aims to preserve the training data information to maintain the performance of GNNs. To reflect this, a natural way is to match the model predictions on the original graph $G$ and its condensed counterpart $G'$. Denote the predictions of an SGC model trained on $G$ as $\hat{\mathbf{Y}}\in\mathbb{R}^{N \times c}$ and those on $G'$ as $\hat{\mathbf{Y}}'\in\mathbb{R}^{N' \times c}$. We have
$\hat{\mathbf{Y}} = \mathbf{F} \mathbf{W}=\mathbf{A}^K \mathbf{X} {\bf W}$ and $\hat{\mathbf{Y}}' = \mathbf{F}' \mathbf{W}'= {\mathbf{A}'}^K \mathbf{X}'{\bf W}' \in \mathbb{R}^{N' \times d}$,

where $K$ denotes the number of SGC layers, $ \mathbf{F}$ and $\mathbf{F}'$ represent the propagated feature matrices on $G$ and $G'$, and $ \mathbf{W}$ and $\mathbf{W}'$ are the weight matrices of the SGC model trained on $G$ and $G'$, respectively. 

Note that feature propagation here is more flexible than the SGC model and is not limited to the formulation \( \mathbf{A}^K \mathbf{X} \); rather, it can be expressed as any linear combination of powers of the adjacency matrix~\cite{maurya2022fsgnn}. Below, we refer to the propagated feature as the node \textit{representations}.  {We first define the distance between two matrices as the L2 norm of their difference. If matrices do not have the same shape, a projection matrix can be applied to transform them into a common shape before computing the distance.}
Then, the following theorem holds:
\begin{theorem}    \label{theo:training_stage}
    The prediction distance in the training stage is bounded by the sum of representation distance and parameter distance.
    \begin{equation}
    \| \mathcal{K}(\hat{\mathbf{Y}}) - \hat{\mathbf{Y}}' \| 
    \leq  \| \mathcal{K}(\mathbf{F}) - \mathbf{F}' \| \cdot \| \mathbf{W}' \| + \| \mathbf{F} \| \cdot \| \mathbf{W} - \mathbf{W}' \|
    \end{equation}
    where $\|\cdot\|$ denotes the L2 norm and $\mathcal{K}(\cdot)$ can be any projection function that aligns the dimensions of $\hat{{\bf Y}}$ and $\hat{{\bf Y}}'$ or ${\bf F}$ and ${\bf F}'$.

\end{theorem}
We provide proof in Appendix~\ref{app:proof_training_stage}. Note that $\|\mathbf{F}\|$ is a constant, and the weight matrix $\| \mathbf{W}' \|$ is naturally constrained due to regularization techniques during model optimization to control its magnitude.  Therefore, Theorem~\ref{theo:training_stage} indicates that by minimizing the representation distance ($\| \mathcal{K}(\mathbf{F}) - \mathbf{F}' \|$) and parameter distance ($\| \mathbf{W} - \mathbf{W}' \|$) between two graphs, the predictions derived from the condensed graph can be close to those of the original graph.

\textbf{Test Stage Objectives.} At the test stage, our condensation goal is to ensure that the GNN model, trained on condensed training data $G'$, generalizes effectively to the test graph, i.e., achieving a low prediction error on the test data. We slightly abuse the notation of $\mathbf{F}$ to represent the propagated feature matrix in the test graph, and denote the test ground truth label as $\mathbf{Y}$ and the predicted test labels from the model trained on $G'$ as $\hat{\mathbf{Y}}''=\mathbf{F}\mathbf{W}'$.  Then we have the following theorem that provides understanding for the test prediction error:

\begin{theorem}   \label{theo:test_stage} 
    The test prediction error of the GNN trained of $G'$ is bounded by the test prediction error of the GNN trained on $G$ plus the parameter distance, as formularized by $ \| \mathbf{Y} - \hat{\mathbf{Y}}'' \| \leq \| \mathbf{Y} - \mathbf{F}\mathbf{W} \| + \|\mathbf{F}\| \cdot \| \mathbf{W}- \mathbf{W}'\|$.

\end{theorem}
We provide proof in Appendix~\ref{app:proof_test_stage}. This inequality incorporates both the original test prediction error and the parameter distance. It indicates that by reducing the parameter distance $\| \mathbf{W}- \mathbf{W}'\|$, the test prediction error becomes more tightly bounded, assuming that the original test prediction error \( \| \mathbf{Y} - \mathbf{F}\mathbf{W}\| \) and the propagated feature matrix \( \mathbf{F} \) remain constant. 

\textbf{Summary - Reframing Objectives for Graph Condensation.} 
Our analysis leads us to a new approach to defining the objectives for GC. Theorems~\ref{theo:training_stage} and~\ref{theo:test_stage} suggest that both training and test stage objectives are upper bounded by the parameter distance $\|\mathbf{W} - \mathbf{W}'\|$. Furthermore, the training stage objectives are additionally upper bounded by the representation distance $dist(\mathbf{F}, \mathbf{F}')$. Thus, we introduce a new GC objective that focuses on minimizing both the \textbf{parameter and representation distances} to optimize the overall performance of the condensed graphs.

\subsection{Efficient Optimization for the New Objective}
Focusing on the new condensation objective we propose, we now discuss its efficient optimization and the establishment of a correspondence between original and condensed nodes to enhance traceability. As our objective includes representation and parameter distance, we separate their discussions as follows. 


\textbf{Optimizing Representation Distance $ \| \mathcal{K}(\hat{\mathbf{Y}}) - \hat{\mathbf{Y}}' \|$.} As discussed in Theorem 1, $\mathcal{K}(\cdot)$ serves as a projection function that aligns the dimensions of $\hat{\bf Y}$ and $\hat{\bf Y}'$. While there are multiple choices of $\mathcal{K}(\cdot)$, we propose to implement it through a linear mapping that assigns each node in the original graph $G$ to one of the synthetic nodes in $G'$. This mapping is formalized as a function $\pi:\{1,\ldots,N\}\to\{1,\ldots N'\}$, represented by the assignment matrix $\mathbf{P} \in \mathbb{R}^{N \times N'}$. The matrix $\mathbf{P}$ is binary, where $\mathbf{P}_{ij} = 1$ if and only if node $i$ in $G$ is assigned to node $j$ in $G'$, i.e., $\pi(i) = j$.  To align $\hat{\mathbf{Y}}$ with $\hat{\mathbf{Y}}'$, we first transform $\hat{\mathbf{Y}}$ to the corresponding dimension using $\mathbf{P}^\top\hat{\mathbf{Y}}$. We then compute the prediction distance as $\|\mathbf{P}^\top \hat{\mathbf{Y}} - \hat{\mathbf{Y}}'\|$, which modifies the minimization of representation distance as  follows:
\begin{equation}
    \begin{aligned}
         \min\nolimits_{\mathbf{F}'} \|\mathcal{K}(\mathbf{F}) - \mathbf{F}'\|
        = \min\nolimits_{\mathbf{P},\mathbf{F}'} \| \mathbf{P}^\top \mathbf{F} - \mathbf{F}' \|.
    \end{aligned}
    \label{eq:representation_distance}
\end{equation}
By solving the above optimization problem, we can derive the assignment matrix ${\bf P}$ and the propagated features ${\bf F}'$ for graph $G'$. In alignment with the popular structure-free GC paradigm~\cite{geom, sfgc, jin2021graph}, we set $\mathbf{F}'$ as the condensed node features and use an identity matrix $\mathbf{I}$ for the condensed adjacency matrix to form a condensed graph that effectively minimizes the representation distance. Notably, the assignment matrix $\mathbf{P}$ and the condensed features $\mathbf{F}'$ can be efficiently derived using any Expectation-Maximization (EM)-based clustering algorithm, such as $k$-means. Since $k$-means inherently minimizes the Sum of Squared Errors (SSE), this optimization aligns with the objective $\min \|\mathbf{P} \mathbf{C} - \mathbf{F}\|$,
where $\mathbf{C}$ represents the cluster centers of feature points $\mathbf{F}$. The procedure iteratively updates the cluster centroids $\mathbf{F}'$ and the assignment matrix $\mathbf{P}$ until convergence is reached. This approach presents two major advantages: (a) \textbf{Traceability}: It provides clear insight into how each original node contributes to the condensed graph, facilitating a better understanding of the condensation process. With traceability, users can manipulate the condensed graph alongside the original one—for example, easily filtering low-quality data when the corresponding original nodes exhibit poor characteristics.
(b) \textbf{Efficiency and Versatility}: Unlike traditional GC methods that need specific GNN models to be trained, this new method does not require any GNN training or gradient calculation and is not limited to a particular GNN architecture, which greatly improves efficiency.

\textbf{Optimizing Parameter Distance $\|{\bf W} - {\bf W}'\|$.} As demonstrated in the above establishment of structure-free condensed graphs, clustering effectively bridges the gap between original and condensed graphs by minimizing representation distance. Within this framework, we further explore how the parameter distance correlates with the assignment matrix ${\bf P}$ used in clustering through the following theorem.

\begin{theorem}
The parameter distance can be bounded by the following inequality:
    \begin{equation}
        \| \mathbf{W} - \mathbf{W}' \| \leq \mathcal{C}\cdot (\max (\text{diag}(\mathbf{\mathbf{P}^\top\mathbf{P}})))^2,
    \label{equ:balance}\end{equation}
    where $\mathcal{C}=\frac{\|\mathbf{F} \| \cdot \| \mathbf{Y}\|({\lambda_{\min}(\mathbf{F}^\top\mathbf{F})}+\|\mathbf{F} \|)}{{\lambda^2_{\min}(\mathbf{F}^\top\mathbf{F})}}$ is a constant, $\mathbf{P}^\top\mathbf{P} \in \mathbb{R}^{N' \times N'}$ is a diagonal matrix with each diagonal entry corresponding to how many original nodes are assigned to each synthetic node. 
\label{theo:bound_of_parameter}
\end{theorem}
We provide proof of Theorem~\ref{theo:bound_of_parameter} in Appendix~\ref{app:proof_bound_of_parameter}. This theorem demonstrates that the parameter distance \( \| \mathbf{W} - \mathbf{W}' \| \) is upper bounded by the maximum elements in $\mathbf{P}^\top\mathbf{P}$.  It suggests that assigning a balanced number of original nodes to each synthetic node can effectively lower the upper bound of the parameter distance. Thus, we plan to integrate this balancing constraint into our condensation process to facilitate the minimization of parameter distance.

\subsection{Graph Condensation via Clustering}\label{sec:clustering}
With clustering established as a powerful and efficient approach for achieving bounded error in GC, we now shift our focus to its practical implementation. To facilitate a clustering-based GC method, we propose the GECC process. We start with the foundational GC paradigm, which provides an efficient method for generating the condensed graph \(G'_t\) from the static original graph \(G_t\) at time step $t$. Building on this, we extend to an evolving graph setting by introducing the incremental condensation paradigm to handle dynamic graph updates efficiently. GECC takes the original graph at time step \( t \) and the condensed node representations from the previous time step \( t-1 \) as input, generating the condensed node representations for the current time step. GECC process can be divided into two main stages: \emph{(i.)} \textbf{Feature Propagation}, and \emph{(ii.)} \textbf{Representation Clustering}. The details of these steps are explained below. 
\subsubsection{Feature Propagation}

The goal of feature propagation is to propagate information from a node's neighbors to provide each node with a richer representation. To enable a GC procedure that eliminates the need for training on the whole graphs, we develop a non-parametric feature propagation module \citep{wu2019simplifying,ma2008bringing,maurya2021improving,hamilton2017inductive} designed to produce node embeddings $\mathbf{F}_t$ in the original graph $G_t$. In our work, by following the propagation method in SGC \citep{wu2019simplifying}, the propagated node representations after \( k \) steps are computed as: $\mathbf{F}_{t,k} = \hat{\mathbf{A}}_t^k \mathbf{X}_t$,
where \( \hat{\mathbf{A}}_t \) is the normalized adjacency matrix of \( G_t \), defined as: $\hat{\mathbf{A}}_t = \tilde{\mathbf{D}}_t^{-\frac{1}{2}} \tilde{\mathbf{A}}_t \tilde{\mathbf{D}}_t^{-\frac{1}{2}}$, with \( \tilde{\mathbf{A}}_t = \mathbf{A}_t + \mathbf{I} \) being the adjacency matrix with self-loops, and \( \tilde{\mathbf{D}}_t \) being the degree matrix of \( \tilde{\mathbf{A}}_t \), where \( (\tilde{\mathbf{D}}_t)_{i,i} = \sum_j \tilde{\mathbf{A}}_{t,i,j} \). To combine information from different propagation steps, we adopt a linearization approach. The final representation for each node is given by:
$\mathbf{F}_t = \sum_{k=0}^{K} \alpha_k \mathbf{F}_{t,k} \label{equ:prop}$.
This formulation ensures that each node's representation captures multi-hop neighborhood information, which fuse the structure and feature information. Note that we allow the weights \(\alpha\) to vary in a large range. In particular, if \(\alpha < 0\), it effectively introduces a ``negative offset'' that can capture heterophilic relationships in the representation~\cite{zhu2021graphheterophily}.

\subsubsection{Representation Clustering}
Once the node representations \(\mathbf{F}_t \in \mathbb{R}^{n_t \times d}\) are obtained, we discard the original graph structure and partition the node representations into groups via the following clustering technique. Each node in the condensed graph \(G'_t\) belongs to one of \(c\) classes. We denote their labels as \(\mathbf{y}'_t \in \{1, 2, \dots, c\}^{n'_t}\) or equivalently via one-hot encoding \(\mathbf{Y}'_t \in \mathbb{R}^{n'_t \times c}\). Following the approach of GCond~\citep{jin2021graph}, we predefine the labels of the condensed nodes so that their class distribution in \(\mathbf{y}'_t\) matches the original distribution in \(\mathbf{y}_t\).

\textit{\underline{A Unified Clustering Formulation.}}
By leveraging the node labels and their feature representations \(\{\mathbf{F}_{t,k}\,\mid\,k=1,\dots,c\}\), we can apply a variety of clustering methods to partition the nodes. Here, we adopt a unified formulation that accommodates both \emph{hard} \(k\)-means and \emph{soft} \(k\)-means. We first split the labeled nodes by their class label \(k\in\{1,2,\dots,c\}\).  
Within class~\(k\), we assign the nodes to \(M_k\) clusters, where \(M_k\) is chosen according to a desired \emph{reduction rate $r$}\footnote{Reduction rate $r$ is defined as (\#nodes in synthetic set)/(\#nodes in training set).}.
Let $M=\sum_{k=1}^{c} M_k=n_{t,k}\times\ r$, where $n_{t,k}$ is number of training nodes in class $k$. 
Next, we formulate the (hard/soft) assignment matrix \(\mathbf{P}_t \in \mathbb{R}^{\,n_t \times M}\).  Each row corresponds to one labeled node, and each column corresponds to \(M\) clusters across all classes.  In more fine-grained notation, if cluster~\(j\) belongs to class~\(k\), then the entry \(\mathbf{P}_{t,i,j}\) indicates how node~\(i\) of class~\(k\) is associated with that cluster. Formally,
\begin{equation}
\mathbf{P}_{t,i,j} \;\in\;
\begin{cases}
\{0,1\},  \text{hard clustering}.\\
[0,1],  \text{\emph{soft} clustering: subject to }\sum_{j}\mathbf{P}_{t,i,j} = 1.
\end{cases}
\end{equation}
In the hard-assignment scenario, each node belongs to exactly one cluster, so \(\mathbf{P}_{t,i,j} = 1\) if node \(i\) is placed in cluster~\(j\) and \(0\) otherwise. In the soft-assignment scenario, each node has fractional memberships across the clusters in its class, with the memberships summing to 1.

Under hard \(k\)-means, each node belongs to exactly one cluster. Under soft \(k\)-means, each node has fractional memberships that sum to 1. We employ the classic algorithms of \(k\)-means~\citep{lloyd1982leastpcm} and fuzzy \(c\)-means~\citep{bezdek1984fcm} in practice.

\textit{\underline{A Balanced SSE Objective.}}
As Theorem~\ref{theo:bound_of_parameter} indicates, assigning a balanced number of original nodes to each synthetic node can effectively reduce the upper bound of the parameter distance, thereby enhancing the objectives in both the training and testing stages. Thus, to obtain balanced clusters, we add a regularization term that penalizes deviations from a uniform cluster size. Concretely, let \(\mathbf{C}_t \in \mathbb{R}^{M \times d}\) collect the centroids for the \(M\) total clusters across all classes. We denote \(\mathbf{1} \in \mathbb{R}^{n_t}\) as the all-ones vector, where \(n_t\) is the total number of labeled nodes. Suppose class \(k\) has \(n_{t,k}\) labeled nodes and is assigned \(M_k\) clusters, so that \(M = \sum_{k=1}^{c} M_k\). To encourage each class to be evenly partitioned among its \(M_k\) clusters, we set
\[
\mathbf{u} \;=\; \Bigl[\,
\underbrace{\tfrac{n_{t,1}}{M_{1}},\,\dots,\,\tfrac{n_{t,1}}{M_{1}}}_{M_{1} \text{ entries}},\ 
\underbrace{\tfrac{n_{t,2}}{M_{2}},\,\dots,\,\tfrac{n_{t,2}}{M_{2}}}_{M_{2} \text{ entries}},\ 
\dots,\ 
\underbrace{\tfrac{n_{t,c}}{M_{c}},\,\dots,\,\tfrac{n_{t,c}}{M_{c}}}_{M_{c} \text{ entries}}
\Bigr]^{\!\top},
\]
so that each block of \(\mathbf{u}\) is constant, corresponding to the perfectly balanced cluster size for each class. 
Since clustering can be stochastic, we repeat it multiple times and choose the best result. The standard metric for clustering quality is the sum of squared errors (SSE). Finally, the balanced SSE objective is then given by:
\begin{equation}
\label{eq:balanced_sse_matrix}
J(\mathbf{P}_t, \mathbf{C}_t)
\;=\;
\bigl\|\mathbf{F}_t - \mathbf{P}_t \mathbf{C}_t \bigr\|^{2}
\;+\;
\,\bigl\|\mathbf{P}_t^\top \mathbf{1} \;-\; \mathbf{u}\bigr\|^{2}.
\end{equation}

\paragraph{Centroid Computation.}
Regardless of whether the assignment is hard or soft, the cluster centroids are computed by:
\begin{equation}
\mathbf{C}_t = \mathbf{D}_{P_t}^{-1} \mathbf{P}_t^\top \mathbf{F}_{t},
\end{equation}
where \(\mathbf{D}_{P_t} \in \mathbb{R}^{n_t \times n_t}\) is a diagonal matrix satisfying $(\mathbf{D}_{P_t})_{k,k}=\sum_{i=1}^{n_t} \mathbf{P}_{t,i,k}$. Finally, as the propagated features implicitly encode structural information, and current SOTA methods~\citep{geom,sfgc} operate on structure-free graphs, we do not explicitly generate new edges. Beyond achieving strong predictive performance, this structure-free design is significantly more efficient than structure-based approaches, which typically require an additional edge-generation process with \(\mathcal{O}((n_tr)^2)\) complexity~\citep{gong2024gc4nc}.

Instead, we use an identity matrix to represent the adjacency among condensed nodes. By combining feature propagation with our representation-based clustering, we form the condensed graph as
$G'_t = \bigl(\mathbf{C}'_{t},\,\mathbf{I}'_{t}\bigr),$
where $\mathbf{I}'_{t}\in \mathbb{R}^{M\times M}$ is an identity matrix. This condensed graph preserves the essential structural and feature information of the original graph \(G_t\), thereby enabling efficient GNN training while substantially reducing computational overhead.

\subsection{Evolving Condensation via Incremental Initialization }
To enable the GECC pipeline to effectively condense graphs that evolve over time and increase in size, we adopt a clustering approach inspired by the K-means++ initialization technique~\citep{arthur2006k}. This method improves the quality of clustering by ensuring a better spread of centroids during initialization. Specifically, for each node class \( c \), we partition the nodes into \( C \) clusters using the following process:

First, let the set of already selected centroids at step \( t-1 \) be \( \mathbf{C}_{t-1} = \{\mathbf{c}_1, \mathbf{c}_2, \dots, \mathbf{c}_k\} \), where \( \mathbf{c}_i \in \mathbb{R}^d \), and let the new coming node set in new time step is $\{\mathbf{x}_1,\mathbf{x}_{2},...,\mathbf{x}_m\}$. For the current step \( t \), we compute \( D(\mathbf{x}_i, \mathbf{C}_{t-1}) \) for each node embedding \( \mathbf{x}_i \in \mathbb{R}^d \) in class \( c \) to consider how well the current centroids represent the data points in the feature space and identify regions of the space that require additional centroids. The distance is defined as the minimum Euclidean distance between \( \mathbf{x}_i \) and any centroid in \( \mathbf{C}_{t-1} \):
\[
D(\mathbf{x}_i, \mathbf{C}_{t-1}) = \min_{\mathbf{c} \in \mathbf{C}_{t-1}} \|\mathbf{x}_i - \mathbf{c}\|^2.
\]
Next, we probabilistically select a new centroid \( \mathbf{c}_{k+1} \) from the set of node embeddings \( \{\mathbf{x}_i\} \) based on the computed distances. The probability of selecting \( \mathbf{x}_i \) is proportional to \( D(\mathbf{x}_i, \mathbf{C}_{t-1})^2 \):
\[
P(\mathbf{x}_i) = \frac{D(\mathbf{x}_i, \mathbf{C}_{t-1})^2}{\sum_{\mathbf{x}_j \in \{\mathbf{x}_1,\mathbf{x}_{2},...,\mathbf{x}_m\}} D(\mathbf{x}_j, \mathbf{C}_{t-1})^2}.
\]
This step ensures that new centroids are more likely to be chosen from areas of the feature space that are underrepresented, thus improving the spread of the centroids. 
In an evolving setting, we expect the condensed graph to grow proportionally with the original graph. Therefore, we sample \(m \times r\) new centroid sets, denoted by \(\{\mathbf{C}_{\Delta}\}\). We then update the set of centroids as follows:
\[
\mathbf{C}_t = \mathbf{C}_{t-1} \cup \{\mathbf{C}_{\Delta}\}.
\]

Finally, after obtaining \( \mathbf{C}_t \), each node embedding \( \mathbf{x}_i \) in class \( c \) is assigned to the nearest centroid:
\[
\text{Cluster}(\mathbf{x}_i) = \arg\min_{\mathbf{c} \in \mathbf{C}_t} \|\mathbf{x}_i - \mathbf{c}\|^2.
\]
This step partitions the nodes into \( C \) clusters, ensuring that each cluster is represented by its respective centroid.

Once the centroids $\mathbf{C}_t$ are obtained through this process, they will be served as the condensed node representations $\mathbf{X}'_t$ in the synthetic graph \( G'_t \) at time step \( t \). By leveraging this probabilistic initialization, the clustering method ensures that new centroids are well-distributed and far from existing ones, facilitating a more effective condensation of graph representations at each time step. The complexity analysis can be found in Appendix~\ref{app:complexity}.

\section{Experiments}
To validate the effectiveness of our proposed GECC, we compare it against classic and SOTA baselines in both non-evolving and evolving scenarios. We first detail the experimental setup, including the construction of benchmark datasets and the experimental settings of each method. Next, we present the node classification performance as a measure of condensation results, alongside a comparison of efficiency. Finally, we empirically demonstrate the effectiveness of the feature aggregation module and the importance of the specific incremental initialization design in our method.  Our code is provided in \url{https://github.com/rockcor/GECC}. 

\subsection{Experimental Setup}

\textbf{Datasets and Baselines.} Following most of the GC papers, we select seven datasets: five transductive datasets, i.e., Citeseer, Cora \citep{kipf2016semi}, Pubmed \citep{namata2012query}, Ogbn-arxiv, and Ogbn-products \citep{hu2020open} and two inductive datasets, Flickr and Reddit \citep{zeng2019graphsaint}. All training graphs are randomly divided into five subsets and each preserving the original class distribution except for Ogbn-arxiv-real, which follows the real timestamps of the publication years. For transductive graphs, nodes in training set are split, whereas inductive graphs are partitioned into subgraphs. Subsequently, the training sets are \textbf{incrementally enlarged}—for example, the first subset forms $G_1$, and the first plus the second subset forms $G_2$.
For additional dataset details, please refer to Appendix~\ref{app:statistics}. We compare GECC with most effective and efficient GC methods, encompassing a diverse range of optimization strategies: (1) gradient matching-based: GCond and GCondX \citep{jin2021graph}; (2) distribution matching-based: GCDM \citep{liu2022graph} and SimGC \citep{xiao2024simple}; (3) trajectory matching-based: GEOM \citep{geom}. We also include node selection baselines (Random~\cite{jin2021graph}, KCenter~\cite{sener2017activekcenter}, Herding~\cite{welling2009herding}) to better compare performance--efficiency trade-offs. The condensed graphs are evaluated using a standard GCN trained for 300 epochs with a learning rate of 0.01, as suggested by GC4NC~\cite{gong2024gc4nc}. The GCN is trained on the synthetic dataset and validated and tested on the original validation and test sets. We also list the results from training a standard GCN in whole dataset for both two settings to show the potential upper bound for graph condensation.

\begin{table*}[ht!]
\centering
\caption{Comparison of different condensation methods in two settings and seven datasets. The evolving setting row show the average test accuracy of five time steps. The best results are in \textbf{bold} and the second-best are \underline{underlined}. "OOM" indicates out-of-memory errors. Average condensation time (seconds) for both settings are listed alongside test accuracy for each method.}
\label{tab:main}
\resizebox{\textwidth}{!}{%
\begin{tabular}{ccccc|cccc|cc|cc|cc|cc|c}
\toprule
\multirow{2}{*}{\textbf{Dataset}} & \multirow{2}{*}{\textbf{Setting}} & \textbf{Random} & \textbf{Herding} & \textbf{KCenter} 
& \multicolumn{2}{c}{\textbf{GCondX}}  
& \multicolumn{2}{c}{\textbf{GCDM}}  
& \multicolumn{2}{c}{\textbf{GCond}}  
& \multicolumn{2}{c}{\textbf{SimGC}}  
& \multicolumn{2}{c}{\textbf{GEOM}}  
& \multicolumn{2}{c}{\textbf{GECC}}  
& {\textbf{Whole}}  \\ 
\cmidrule(l){3-5}\cmidrule(l){6-7}\cmidrule(l){8-9}\cmidrule(l){10-11}\cmidrule(l){12-13}\cmidrule(l){14-15}\cmidrule(l){16-17}\cmidrule(l){18-18}
& & Acc. $\uparrow$ & Acc. $\uparrow$& Acc. $\uparrow$& Acc. $\uparrow$ & Time $\downarrow$ & Acc. $\uparrow$ & Time $\downarrow$ & Acc. $\uparrow$ & Time $\downarrow$ & Acc. $\uparrow$ & Time $\downarrow$ & Acc. $\uparrow$ & Time $\downarrow$ & Acc. $\uparrow$ & Time $\downarrow$ & Acc. $\uparrow$ \\
\midrule

\multirow{2}{*}{\textit{Citeseer}} 
& Non-Evolving
  & 62.62 & 66.66 & 59.04
  & 68.38 & \multirow{2}{*}{\textit{506}}
  & 69.35 & \multirow{2}{*}{\textit{654}}
  & 72.08 & \multirow{2}{*}{\textit{218}}
  & 66.40 & \multirow{2}{*}{\textit{1680}}
  & {\underline{73.03}} & \multirow{2}{*}{\textit{1362}}
  & {\textbf{73.25}} & \multirow{2}{*}{\textcolor{red}{\textbf{\textit{1.7}}}}
  & 72.11
\\
& Evolving
  & 50.65 & 53.47 & 47.99
  & 50.85 &
  & 60.51 &
  & {\underline{61.51}} &
  & 57.42 &
  & 58.95 &
  & {\textbf{65.48}} &
  & 63.57
\\ 
\cmidrule{1-18}

\multirow{2}{*}{\textit{Cora}}
& Non-Evolving
  & 72.24 & 73.77 & 70.55
  & 78.60 & \multirow{2}{*}{\textit{332}}
  & 80.54 & \multirow{2}{*}{\textit{1190}}
  & 80.68 & \multirow{2}{*}{\textit{143}}
  & 79.60 & \multirow{2}{*}{\textit{1644}}
  & {\underline{82.82}} & \multirow{2}{*}{\textit{1331}}
  & {\textbf{82.99}} & \multirow{2}{*}{\textcolor{red}{\textbf{\textit{1.7}}}}
  & 81.23
\\
& Evolving
  & 58.00 & 63.07 & 59.90
  & 67.18 &
  & {\underline{77.14}} &
  & 74.54 &
  & 64.42 &
  & 72.56 &
  & {\textbf{77.36}} &
  & 76.34
\\ 
\cmidrule{1-18}

\multirow{2}{*}{\textit{Pubmed}}
& Non-Evolving
  & 71.84 & 75.53 & 74.00
  & 71.97 & \multirow{2}{*}{\textit{247}}
  & 76.46 & \multirow{2}{*}{\textit{502}}
  & 77.48 & \multirow{2}{*}{\textit{311}}
  & 76.80 & \multirow{2}{*}{\textit{1654}}
  & {\underline{78.49}} & \multirow{2}{*}{\textit{995}}
  & {\textbf{80.24}} & \multirow{2}{*}{\textcolor{red}{\textbf{\textit{1.4}}}}
  & 78.65
\\
& Evolving
  & 66.37 & 66.31 & 64.38
  & 62.65 &
  & 74.26 &
  & {\underline{74.49}} &
  & 71.38 &
  & 70.25 &
  & {\textbf{76.74}} &
  & 76.18
\\ 
\cmidrule{1-18}

\multirow{2}{*}{\textit{Flickr}}
& Non-Evolving
  & 44.68 & 45.12 & 43.53
  & 46.58 & \multirow{2}{*}{\textit{610}}
  & {\textbf{46.99}} & \multirow{2}{*}{\textit{1447}}
  & 45.88 & \multirow{2}{*}{\textit{354}}
  & 41.01 & \multirow{2}{*}{\textit{7487}}
  & 46.13 & \multirow{2}{*}{\textit{758}}
  & {\underline{46.63}} & \multirow{2}{*}{\textcolor{red}{\textbf{\textit{7.1}}}}
  & 47.53
\\
& Evolving
  & 44.70 & 44.66 & 44.33
  & {\underline{45.63}} &
  & 45.52 &
  & 44.98 &
  & 41.94 &
  & 45.43 &
  & {\textbf{45.78}} &
  & 46.97
\\ 
\cmidrule{1-18}

\multirow{2}{*}{\textit{Ogbn-arxiv}}
& Non-Evolving
  & 60.19 & 57.70 & 58.66
  & 59.93 & \multirow{2}{*}{\textit{2895}}
  & 64.23 & \multirow{2}{*}{\textit{3020}}
  & 60.71 & \multirow{2}{*}{\textit{686}}
  & 65.26 & \multirow{2}{*}{\textit{2687}}
  & {\textbf{69.59}} & \multirow{2}{*}{\textit{1685}}
  & {\underline{66.71}} & \multirow{2}{*}{\textcolor{red}{\textbf{\textit{10}}}}
  & 70.95
\\
& Evolving
  & 56.04 & 57.57 & 56.21
  & 60.73 &
  & 62.50 &
  & 59.98 &
  & 64.97 &
  & {\textbf{66.30}} &
  & {\underline{65.42}} &
  & 70.40
\\ 
\cmidrule{1-18}
\multirow{2}{*}{\textit{Ogbn-arxiv-real}}
& Non-Evolving
  & 60.19 & 57.70 & 58.66
  & 59.93 & \multirow{2}{*}{\textit{1328}}
  & 64.23 & \multirow{2}{*}{\textit{1862}}
  & 60.71 & \multirow{2}{*}{\textit{617}}
  & 65.26 & \multirow{2}{*}{\textit{3240}}
  & {\textbf{69.59}} & \multirow{2}{*}{\textit{1646}}
  & {\underline{66.71}} & \multirow{2}{*}{\textcolor{red}{\textbf{\textit{8.6}}}}
  & 70.95
\\
& Evolving
  & 56.04 & 57.57 & 56.21
  & 59.33 &
  & 61.36 &
  & 59.46 &
  & 59.34 &
  & {\underline{58.28}} &
  & {\textbf{61.48}} &
  & 70.40
\\ 
\cmidrule{1-18}
\multirow{2}{*}{\textit{Ogbn-products}}
& Non-Evolving
  & 60.19 & 57.70 & 58.66
  & OOM & \multirow{2}{*}{\centering -}
  & OOM & \multirow{2}{*}{\centering -}
  & OOM & \multirow{2}{*}{\centering -}
  & {\underline{61.71}} & \multirow{2}{*}{\centering \textit{71489}}
  & OOM & \multirow{2}{*}{\centering -}
  & {\textbf{66.32}} & \multirow{2}{*}{\centering \textcolor{red}{\textbf{\textit{147}}}}
  & 73.40
\\
& Evolving
  & 41.36 & 44.26 & 38.93
  & OOM &
  & OOM &
  & OOM &
  & {\underline{61.93}} &
  & OOM &
  & {\textbf{64.03}} &
  & 73.88
\\ \midrule
\multirow{2}{*}{\textit{Reddit}}
& Non-Evolving
  & 55.73 & 59.34 & 48.28
  & 88.25 & \multirow{2}{*}{\textit{2673}}
  & 89.82 & \multirow{2}{*}{\textit{6130}}
  & 89.96 & \multirow{2}{*}{\textit{337}}
  & 90.78 & \multirow{2}{*}{\textit{6611}}
  & {\underline{91.33}} & \multirow{2}{*}{\textit{1816}}
  & {\textbf{91.37}} & \multirow{2}{*}{\textcolor{red}{\textbf{\textit{4.9}}}}
  & 93.70
\\
& Evolving
  & 51.31 & 48.94 & 48.53
  & 79.02 &
  & 87.93 &
  & 82.68 &
  & {\underline{89.85}} &
  & 67.91 &
  & {\textbf{90.02}} &
  & 93.92
\\
\bottomrule
\end{tabular}}
\end{table*}

\begin{figure*}[ht!]
        \centering
        \includegraphics[width=\linewidth]{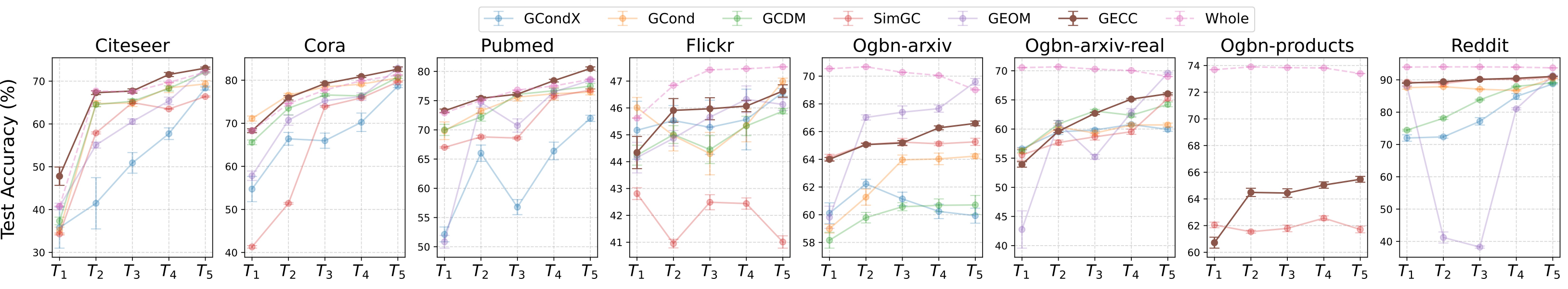}
    \caption{Comparison of test accuracy of different GC methods across five time steps.}
    \label{fig:evolving}
\end{figure*}
\textbf{Implementation Details.} \label{sec:hyper}
To ensure a fair reproduction and comparison of baseline methods, we use the best hyperparameters reported in their original papers. For intermediate evaluation, we follow the GC4NC benchmark~\cite{gong2024gc4nc}, which restricts the number of evaluations to 10 during the whole condensation process. 
All baselines adopt the training from scratch strategy in evolving stetting, i.e., do not reuse the previous condensed graphs.
For the hyperparameters of our method, we tune them within a limited range, specifically $\alpha_0, \alpha_1, \alpha_2 \in [-0.3,0.9]$ with 0.1 interval. We introduce a negative offset to capture heterophilious properties in graphs~\cite{zhu2021graphheterophily,gong2023neighborhood}. Following prior work, we fix the maximum propagation depth in Equation~\ref{equ:prop} as $K=2$. 
In addition, learning rate, epochs and dropout of downstream GCN are all fixed as 0.01, 300 and 0.5, except the weight decay is selected from
$\{0.001, 0.0005\}$. We use soft clustering for small datasets and the repeat times are set to 50. The fuzziness are selected from $\{1.0,\; 1.1,\; 1.3\}$, respectively. For large datasets, we run standard hard $k$-means only one time. The maximum number of iterations and the $k$-means threshold are set to 300 and $1 \times 10^{-8}$, respectively.  All experiments are run ten times then we report the average. See more details in Appendix~\ref{app:exp}.

\begin{table*}[t!]\label{tab:transfer}
\centering
\caption{Transferability: GECC is compatible with various GNN architectures. ``Average'' denotes the mean $\pm$ standard deviation of test accuracy across different GNN models.}
\label{tabs:transfer}
\resizebox{\linewidth}{!}{
\begin{tabular}{lccccccccc}
\toprule
\multirow{2}{*}{\bfseries Model} & \multicolumn{3}{c}{\textbf{Cora ($r=0.5$)}} & \multicolumn{3}{c}{\textbf{Ogbn-arxiv ($r=0.01$)}} & \multicolumn{3}{c}{\textbf{Reddit ($r=0.001$)}} \\
\cmidrule(lr){2-4} \cmidrule(lr){5-7} \cmidrule(lr){8-10}
                & GCond   & GEOM   & GECC   & GCond    & GEOM   & GECC   & GCond    & GEOM   & GECC \\
\midrule
GCN           & 81.69   & 83.06  & 82.71  & 65.83    & 69.49  & 66.60  & 88.01    & 91.57  & 91.28  \\
SGC           & 81.38   & 83.15  & 82.67  & 64.76    & 67.51  & 65.55  & 88.69    & 90.58  & 92.10  \\
APPNP         & 80.80   & 83.33  & 81.97  & 64.97    & 67.03  & 64.45  & 85.90    & 89.31  & 91.07  \\
Cheby         & 77.61   & 80.49  & 78.45  & 60.52    & 62.35  & 60.17  & 74.51    & 82.58  & 82.26  \\
GraphSage     & 77.52   & 74.12  & 78.45  & 60.38    & 62.19  & 60.24  & 74.47    & 82.60  & 82.24  \\
GAT           & 76.47   & 81.85  & 82.09  & 61.90    & 64.70  & 64.35  & 88.84    & 89.19  & 90.40  \\ \midrule
Average       & 79.25$\pm$1.76   & \underline{81.00$\pm$3.09}   & \bfseries 81.06$\pm$2.01   & 63.06$\pm$2.23   & \bfseries 65.55$\pm$2.74   & \underline{63.56$\pm$2.31}   & 83.40$\pm$5.41   & \underline{87.64$\pm$3.77}   & \bfseries 88.23$\pm$3.28   \\
\bottomrule
\end{tabular}}
\end{table*}

\subsection{Performance and Efficiency Comparison in the non-Evolving and Evolving Setting}

\subsubsection{Performance Comparison}
To compare the effectiveness of GECC with the baselines, we utilize the generated condensed graph data to train a standard GCN, reporting both test accuracies and standard deviations in Table \ref{tab:main}. For each dataset, we report two test accuracies: one for the \textit{Non-Evolving} setting, reflecting the accuracy achieved by training the GCN on the graph at the final time step ($t = 5$), corresponding to the largest graph size; and one for the \textit{Evolving} setting, representing the average accuracy across all five time steps as the graph evolves. It is also noteworthy that, to evaluate our proposed method on real-world dynamic datasets such as \textit{Ogbn-Arxiv-real}, we utilize the original metadata from \textit{Ogbn-arxiv}~\citep{hu2020open} and use node timestamps to divide the data into five sequential time steps.

Additionally, Figure \ref{fig:evolving} illustrates the test accuracy across different time steps, which compares the GECC evolving capability with the existing baselines. Our analysis reveals several key insights: 

\textbf{Non-evolving setting --} GECC outperforms the baselines on almost all datasets, including the \emph{whole} dataset performance for some datasets, which highlights the effectiveness of our training-free approach in the non-evolving setting. The only exceptions are observed in the \textit{Flickr} and \textit{Ogbn-arxiv} datasets, where GECC ranks second by a minimal margin. In addition, unlike other baselines, GECC does not fail on extremely large datasets like \textit{Ogbn-products} due to memory constraints. 

\textbf{Evolving setting --} By analyzing Table ~\ref{tab:main}, we observe that GECC consistently outperforms existing baselines across almost all datasets, often by a large margin. The only exception is \textit{Ogbn-arxiv}, where GECC ranks second. ALso, Figure~\ref{fig:evolving} highlights that GECC achieves the best test accuracy at early time steps. For instance, on \textit{Ogbn-arxiv}, it surpasses 65\% accuracy by the second time step, a large gap compared to existing baselines. This demonstrates GECC’s efficiency in leveraging limited data for superior generalization, whereas other methods struggle to reach comparable performance early because they need sufficient data for training. For the \textit{Ogbn-arxiv-real} dataset, a real-world graph with inherent dynamics based on original timestamps, GECC still achieves the best evolving performance, indicating that our method remains effective in realistic dynamic scenarios. 
\textbf{Moreover}, we observe that GECC surpasses the performance achieved on the whole dataset for relatively smaller graphs such as \textit{Cora}, \textit{Citeseer}, and \textit{Pubmed}. This result highlights the effectiveness of our training-free condensation approach in achieving a "lossless" objective during graph evolution. \textbf{Additionally}, Figure~\ref{fig:evolving} illustrates the robustness of GECC in steadily improving performance as training size increases. This trend fails to hold for baselines, particularly for GEOM on \textit{Reddit}. We conjecture that the trajectory matching method heavily relies on the performance of the pre-trained GNN, making it susceptible to the twofold influence of data size changes: first, during the pre-training stage, and second, during the condensation stage.


\vskip -1.5em
\subsubsection{Efficiency Comparison}
Comparing GECC with the baselines in Figure~\ref{fig:accuracy_vs_time}, it is evident that GECC exhibits superior efficiency and scalability. It maintains stable performance while effectively managing computational resources as the graph evolves. Numerical experiments show sublinear runtime growth: a fitted power-law exponent ($\approx 0.3$) indicates reasonably good scalability. Although certain model-based GC methods such as GEOM may slightly outperform GECC at specific time steps, GECC achieves over 100 times faster condensation time and demonstrates a significantly slower increase in computational overhead. For results in more datasets, please refer to Appendix~\ref{app:exp}.


\begin{figure}[ht!]
    \centering
    \includegraphics[width=\linewidth]{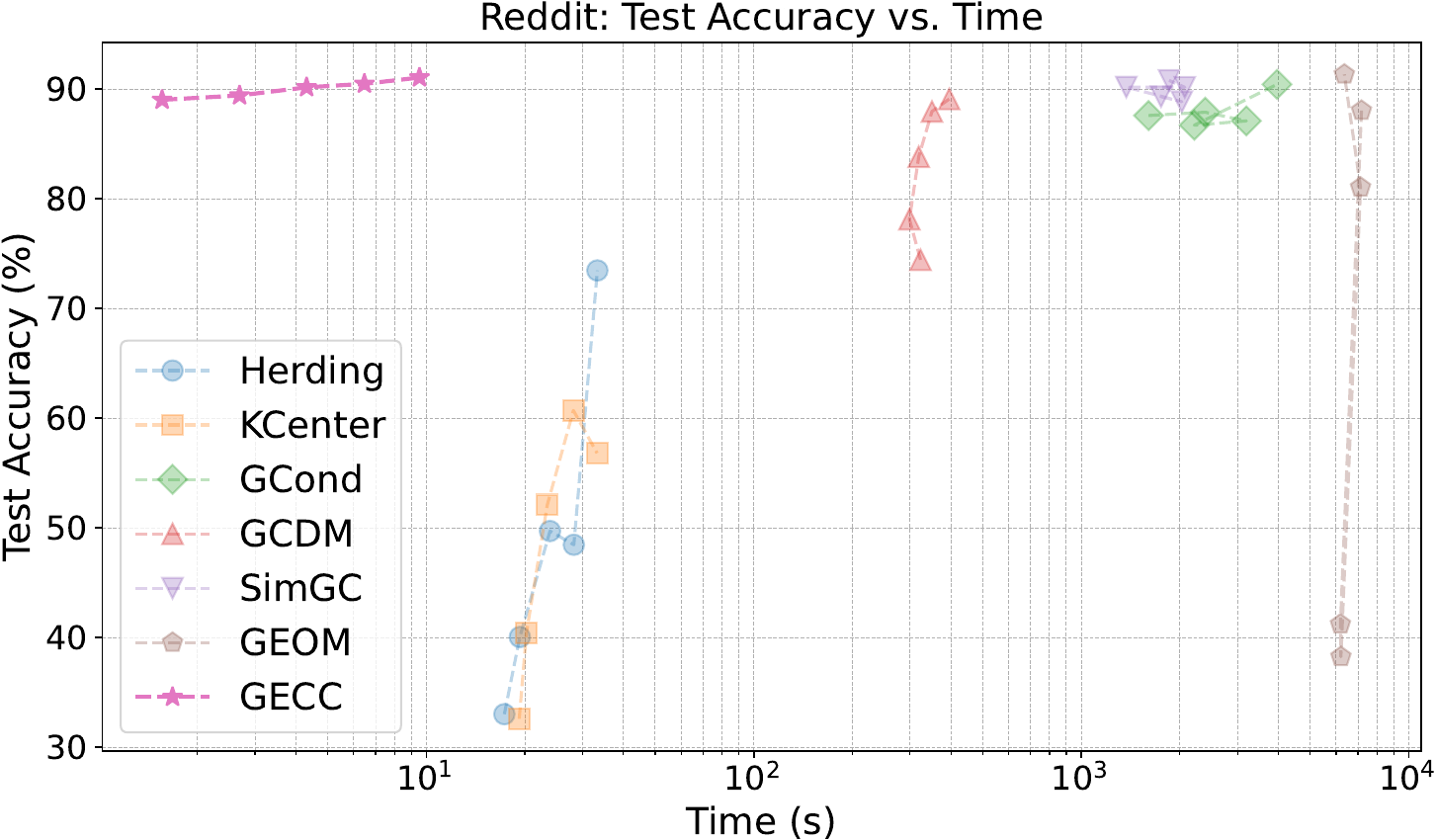}
    \vskip -1em
    \caption{Test accuracy vs. condensation time on the \textit{Reddit} dataset (top-left is better).}
    \label{fig:accuracy_vs_time}
\end{figure}

\subsubsection{Transferability}
A crucial factor in evaluating GC methods is determining whether the condensed data can effectively train various GNNs from a data-centric perspective. Unlike GECC, which adopts a training-free condensation approach, most existing methods generate condensed graphs that are inherently dependent on the backbone GNN used during condensation, such as GCN \citep{hashemi2024comprehensive,gong2024gc4nc}. This reliance can introduce inductive biases, potentially hindering their adaptability to other GNN architectures. 

Table~\ref{tabs:transfer} shows that the condensed graphs generated by \textsc{GECC} demonstrate robust generalization across diverse architectures when compared to other SOTA baselines. Even in \textbf{Ogbn-arxiv}, \textsc{GECC} outperforms \textsc{GEOM} w.r.t consistency, as indicated by its lower standard deviation across downstream GNN models, highlighting the advantage of \textsc{GECC}'s model-agnostic design.

\subsection{Ablation Studies}
To investigate the effectiveness of each module of our method, we conduct the following ablation studies to study the impact of feature propagation, incremental $k$-means++, and balanced SSE score. 
\subsubsection{Impact of Feature Propagation}
As feature propagation (Equation~\ref{equ:prop}) is crucial for generating informative node representations, we compare \textsc{GECC} against an ablated version without feature propagation (labeled \emph{w/o propagation}) across all time steps. Specifically, for \emph{w/o propagation}, we set \(\alpha_{0}=1\) and all other \(\alpha\) coefficients to \(0\), keeping all other hyperparameters identical. Each experiment is repeated ten times, and we report the average results.

\begin{figure*}[t!]
    \centering
    \includegraphics[width=0.8\linewidth]{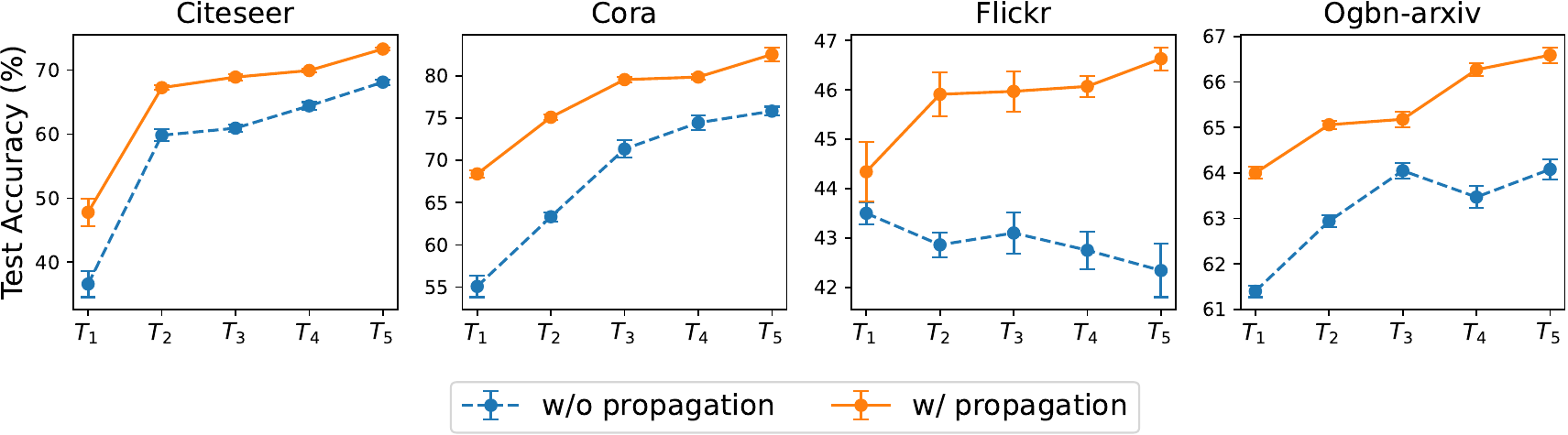}
    \caption{Performance comparison between \textsc{GECC} and \textsc{GECC} without feature aggregation.}
    \label{fig:abla-prop}
\end{figure*}

Figure~\ref{fig:abla-prop} illustrates that \textsc{GECC} with feature propagation outperforms the version without propagation. Notably, in some datasets (e.g., Flickr), the \emph{w/o propagation} approach exhibits a downward trend even as more graph data is introduced, suggesting that the raw node features may contain substantial noise. In contrast, \textsc{GECC} with feature propagation steadily improves as the dataset size increases, highlighting that propagated features effectively mitigate noise and bolster clustering performance.

\subsubsection{Impact of Incremental \(k\)-Means++}
We evaluate the effect of reusing previous condensed results when clustering at each time step. Specifically, we compare \textsc{GECC} against a variant that does \emph{not} reuse prior centroids---denoted \textit{w/o incremental \(k\)-means++}---and instead initializes centroids from scratch via a standard \(k\)-means++ procedure at every time step.
\begin{figure*}[t!]
    \centering
    \includegraphics[width=0.8\linewidth]{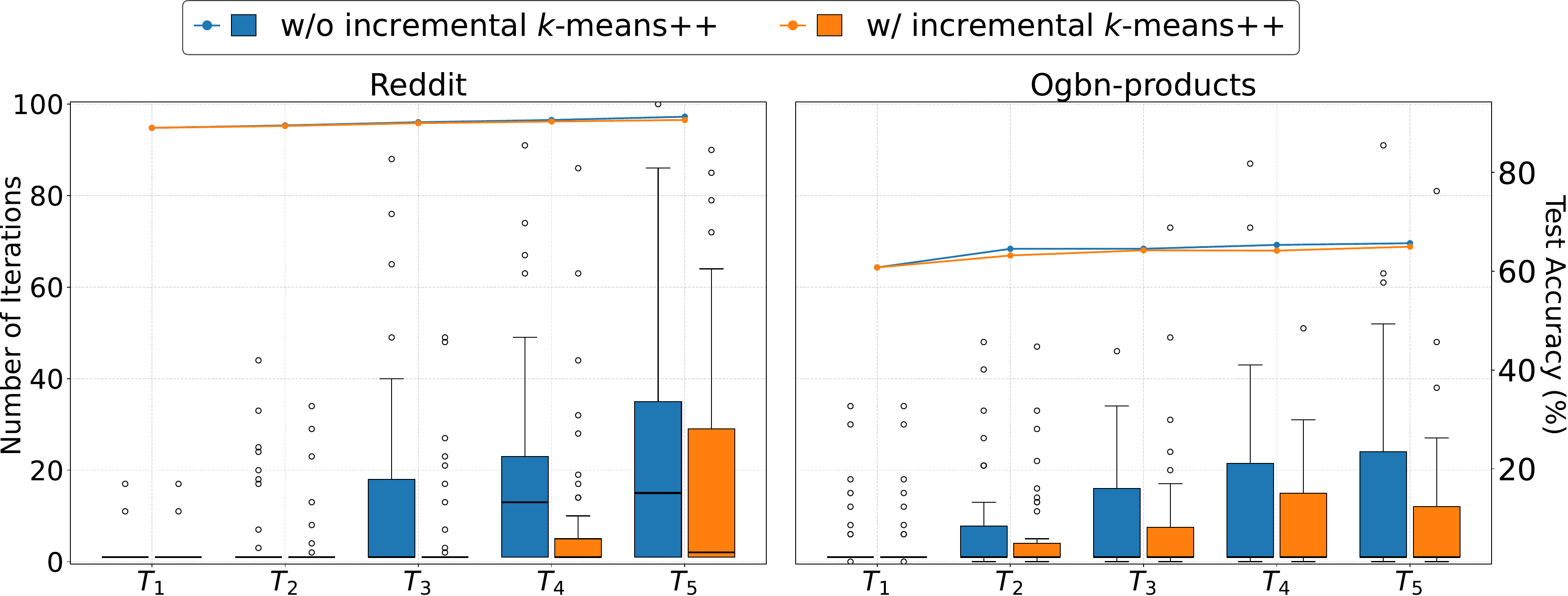}
    \caption{Ablation study. Boxplots show the mean and quartiles of the number of iterations required for clustering.}
    \label{fig:abla-incrementalkmeans++}
\end{figure*}

As shown in Figure~\ref{fig:abla-incrementalkmeans++}, reusing previously learned cluster centers via incremental \(k\)-means++ significantly reduces the required number of iterations for convergence, especially as the graph grows larger. For example, on the \textit{Reddit} dataset at time step~5, incremental initialization only requires about 10\% of the iterations needed when initializing from scratch. We omit results on smaller datasets because they typically converge in fewer than 10 iterations.


\subsubsection{Relation between GC and clustering objective}
The balanced Sum of Squared Errors (SSE) is a key contribution derived from our theoretical analysis. To assess its impact, we perform an ablation study comparing the performance of GECC with and without repetitive clustering. As shown in Table~\ref{tab:comparison_sse}, for instance, on the \textit{Citeseer} dataset, applying repetitive clustering to select the lowest SSE leads to an absolute performance improvement of 2.7\%. These results show that a lower SSE correlates with higher test accuracy.
\begin{table}[t!]
\centering
\caption{Comparison of average test accuracy (\%) and balanced SSE on three benchmark datasets across five time steps. ``Bal.\ SSE'' indicates the balanced SSE.}
\resizebox{\linewidth}{!}{%
\begin{tabular}{lcccccc}
\toprule
& \multicolumn{2}{c}{\textit{Citeseer}} & \multicolumn{2}{c}{\textit{Cora}} & \multicolumn{2}{c}{\textit{Pubmed}} \\
\cmidrule(r){2-3} \cmidrule(r){4-5} \cmidrule(r){6-7}
& Acc.~(\(\uparrow\)) & Bal.\ SSE~(\(\downarrow\)) 
& Acc.~(\(\uparrow\)) & Bal.\ SSE~(\(\downarrow\)) 
& Acc.~(\(\uparrow\)) & Bal.\ SSE~(\(\downarrow\)) \\
\midrule
w/ rep.\ clustering  & 65.45 & 2.39 & 77.08 & 4.52 & 76.32 & 6.65 \\
w/o rep.\ clustering & 63.75 & 9.83 & 74.82 & 9.47 & 75.99 & 9.89 \\
\bottomrule
\end{tabular}}
\label{tab:comparison_sse}
\end{table}

\section{Conclusion and Outlook}
In this study, we address the challenge of evolving graph condensation. We observe that a universal clustering framework can naturally optimize the assignment matrix, thereby achieving the common objectives of existing GC methods. Additionally, we propose a novel \emph{balanced SSE} metric that further tightens the upper bound of these objectives. In the evolving setting, we find that our clustering approach can be readily adapted to an incremental version, termed \emph{incremental \(k\)-means++}. 

Experimental results demonstrate that balanced SSE improves the performance of clustering-based GC, and incremental \(k\)-means++ significantly reduces the number of iterations, thereby enhancing efficiency in evolving environments. Future work includes developing more efficient and scalable clustering techniques, especially soft clustering algorithms for larger graph datasets and adaptively optimizing multi-hop weights, which could be beneficial when the graph keeps evolving over time.

\section{Acknowledgement}
This research was supported by the U.S. National Science Foundation under Award No. 2504088.
This research was also partially supported by the US National Science Foundation under Award Numbers 2319449, 2312502, and 2442172, as well as the US National Institute of Diabetes and Digestive and Kidney Diseases of the US National Institutes of Health under Award Number K25DK135913.

\clearpage
\bibliographystyle{ACM-Reference-Format}
\balance
\bibliography{0main}

@article{hashemi2024comprehensive,
  title={A comprehensive survey on graph reduction: Sparsification, coarsening, and condensation},
  author={Hashemi, Mohammad and Gong, Shengbo and Ni, Juntong and Fan, Wenqi and Prakash, B Aditya and Jin, Wei},
  journal={IJCAI},
  year={2024}
}

@article{wu2020comprehensive,
  title={A comprehensive survey on graph neural networks},
  author={Wu, Zonghan and Pan, Shirui and Chen, Fengwen and Long, Guodong and Zhang, Chengqi and Philip, S Yu},
  journal={IEEE transactions on neural networks and learning systems},
  volume={32},
  number={1},
  pages={4--24},
  year={2020},
  publisher={IEEE}
}

@inproceedings{pyg,
  title={Fast Graph Representation Learning with {PyTorch Geometric}},
  author={Fey, Matthias and Lenssen, Jan E.},
  booktitle={ICLR Workshop on Representation Learning on Graphs and Manifolds},
  year={2019},
}

@inproceedings{su2023robustincremental,
  title={Towards robust graph incremental learning on evolving graphs},
  author={Su, Junwei and Zou, Difan and Zhang, Zijun and Wu, Chuan},
  booktitle={International Conference on Machine Learning},
  pages={32728--32748},
  year={2023},
  organization={PMLR}
}

@inproceedings{geom,
  title={Navigating Complexity: Toward Lossless Graph Condensation via Expanding Window Matching},
  author={Zhang, Yuchen and Zhang, Tianle and Wang, Kai and Guo, Ziyao and Liang, Yuxuan and Bresson, Xavier and Jin, Wei and You, Yang},
  booktitle={International Conference on Machine Learning},
  pages={60379--60395},
  year={2024},
  organization={PMLR}
}

@article{bezdek1984fcm,
  title={FCM: The fuzzy c-means clustering algorithm},
  author={Bezdek, James C and Ehrlich, Robert and Full, William},
  journal={Computers \& geosciences},
  volume={10},
  number={2-3},
  pages={191--203},
  year={1984},
  publisher={Elsevier}
}

@inproceedings{welling2009herding,
  title={Herding dynamical weights to learn},
  author={Welling, Max},
  booktitle={Proceedings of the 26th annual international conference on machine learning},
  pages={1121--1128},
  year={2009}
}

@article{numpy,
 title         = {Array programming with {NumPy}},
 author        = {Charles R. Harris and K. Jarrod Millman and St{\'{e}}fan J.
                 van der Walt and Ralf Gommers and Pauli Virtanen and David
                 Cournapeau and Eric Wieser and Julian Taylor and Sebastian
                 Berg and Nathaniel J. Smith and Robert Kern and Matti Picus
                 and Stephan Hoyer and Marten H. van Kerkwijk and Matthew
                 Brett and Allan Haldane and Jaime Fern{\'{a}}ndez del
                 R{\'{i}}o and Mark Wiebe and Pearu Peterson and Pierre
                 G{\'{e}}rard-Marchant and Kevin Sheppard and Tyler Reddy and
                 Warren Weckesser and Hameer Abbasi and Christoph Gohlke and
                 Travis E. Oliphant},
 year          = {2020},
 month         = sep,
 journal       = {Nature},
 volume        = {585},
 number        = {7825},
 pages         = {357--362},
 doi           = {10.1038/s41586-020-2649-2},
 publisher     = {Springer Science and Business Media {LLC}},
 url           = {https://doi.org/10.1038/s41586-020-2649-2}
}

@article{sener2017activekcenter,
  title={Active learning for convolutional neural networks: A core-set approach},
  author={Sener, Ozan and Savarese, Silvio},
  journal={arXiv preprint arXiv:1708.00489},
  year={2017}
}

@article{lloyd1982leastpcm,
  title={Least squares quantization in PCM},
  author={Lloyd, Stuart},
  journal={IEEE transactions on information theory},
  volume={28},
  number={2},
  pages={129--137},
  year={1982},
  publisher={IEEE}
}

@inproceedings{zhu2021graphheterophily,
  title={Graph neural networks with heterophily},
  author={Zhu, Jiong and Rossi, Ryan A and Rao, Anup and Mai, Tung and Lipka, Nedim and Ahmed, Nesreen K and Koutra, Danai},
  booktitle={Proceedings of the AAAI conference on artificial intelligence},
  volume={35},
  number={12},
  pages={11168--11176},
  year={2021}
}

@inproceedings{gong2023neighborhood,
  title={Neighborhood homophily-based graph convolutional network},
  author={Gong, Shengbo and Zhou, Jiajun and Xie, Chenxuan and Xuan, Qi},
  booktitle={Proceedings of the 32nd ACM International Conference on Information and Knowledge Management},
  pages={3908--3912},
  year={2023}
}

@article{hu2021ogb,
  title={Ogb-lsc: A large-scale challenge for machine learning on graphs},
  author={Hu, Weihua and Fey, Matthias and Ren, Hongyu and Nakata, Maho and Dong, Yuxiao and Leskovec, Jure},
  journal={arXiv preprint arXiv:2103.09430},
  year={2021}
}

@article{maurya2022fsgnn,
  title={Simplifying approach to node classification in graph neural networks},
  author={Maurya, Sunil Kumar and Liu, Xin and Murata, Tsuyoshi},
  journal={Journal of Computational Science},
  volume={62},
  pages={101695},
  year={2022},
  publisher={Elsevier}
}

@inproceedings{complexity,
  title={Understanding and bridging the gaps in current GNN performance optimizations},
  author={Huang, Kezhao and Zhai, Jidong and Zheng, Zhen and Yi, Youngmin and Shen, Xipeng},
  booktitle={Proceedings of the 26th ACM SIGPLAN Symposium on Principles and Practice of Parallel Programming},
  pages={119--132},
  year={2021}
}

@inproceedings{liu2024review,
  title={A review of graph neural networks in epidemic modeling},
  author={Liu, Zewen and Wan, Guancheng and Prakash, B Aditya and Lau, Max SY and Jin, Wei},
  booktitle={Proceedings of the 30th ACM SIGKDD Conference on Knowledge Discovery and Data Mining},
  pages={6577--6587},
  year={2024}
}

@inproceedings{fan2019graph,
  title={Graph neural networks for social recommendation},
  author={Fan, Wenqi and Ma, Yao and Li, Qing and He, Yuan and Zhao, Eric and Tang, Jiliang and Yin, Dawei},
  booktitle={The world wide web conference},
  pages={417--426},
  year={2019}
}

@inproceedings{gcsntk,
  title={Fast graph condensation with structure-based neural tangent kernel},
  author={Wang, Lin and Fan, Wenqi and Li, Jiatong and Ma, Yao and Li, Qing},
  booktitle={Proceedings of the ACM on Web Conference 2024},
  pages={4439--4448},
  year={2024}
}

@article{jin2021graph,
  title={Graph condensation for graph neural networks},
  author={Jin, Wei and Zhao, Lingxiao and Zhang, Shichang and Liu, Yozen and Tang, Jiliang and Shah, Neil},
  journal={arXiv preprint arXiv:2110.07580},
  year={2021}
}

@article{gong2024gc4nc,
  title={GC4NC: A Benchmark Framework for Graph Condensation on Node Classification with New Insights},
  author={Gong, Shengbo and Ni, Juntong and Sachdeva, Noveen and Yang, Carl and Jin, Wei},
  journal={arXiv preprint arXiv:2406.16715},
  year={2024}
}

@inproceedings{jin2022condensing,
  title={Condensing graphs via one-step gradient matching},
  author={Jin, Wei and Tang, Xianfeng and Jiang, Haoming and Li, Zheng and Zhang, Danqing and Tang, Jiliang and Yin, Bing},
  booktitle={Proceedings of the 28th ACM SIGKDD Conference on Knowledge Discovery and Data Mining},
  pages={720--730},
  year={2022}
}

@article{liu2022graph,
  title={Graph condensation via receptive field distribution matching},
  author={Liu, Mengyang and Li, Shanchuan and Chen, Xinshi and Song, Le},
  journal={arXiv preprint arXiv:2206.13697},
  year={2022}
}

@article{sun2024gc,
  title={Gc-bench: An open and unified benchmark for graph condensation},
  author={Sun, Qingyun and Chen, Ziying and Yang, Beining and Ji, Cheng and Fu, Xingcheng and Zhou, Sheng and Peng, Hao and Li, Jianxin and Yu, Philip S},
  journal={arXiv preprint arXiv:2407.00615},
  year={2024}
}

@article{du2019gntk,
  title={Graph neural tangent kernel: Fusing graph neural networks with graph kernels},
  author={Du, Simon S and Hou, Kangcheng and Salakhutdinov, Russ R and Poczos, Barnabas and Wang, Ruosong and Xu, Keyulu},
  journal={Advances in neural information processing systems},
  volume={32},
  year={2019}
}

@journal{
li2025a,
title={A Precompute-Then-Adapt Approach for Efficient Graph Condensation},
author={Yuan Li and Jun Hu and Zemin Liu and Bryan Hooi and Jia Chen and Bingsheng He},
year={2025},
url={https://openreview.net/forum?id=kwagvI8Anf}
}

@inproceedings{wu2019simplifying,
  title={Simplifying graph convolutional networks},
  author={Wu, Felix and Souza, Amauri and Zhang, Tianyi and Fifty, Christopher and Yu, Tao and Weinberger, Kilian},
  booktitle={International conference on machine learning},
  pages={6861--6871},
  year={2019},
  organization={PMLR}
}

@article{ma2008bringing,
  title={Bringing PageRank to the citation analysis},
  author={Ma, Nan and Guan, Jiancheng and Zhao, Yi},
  journal={Information Processing \& Management},
  volume={44},
  number={2},
  pages={800--810},
  year={2008},
  publisher={Elsevier}
}

@article{maurya2021improving,
  title={Improving graph neural networks with simple architecture design},
  author={Maurya, Sunil Kumar and Liu, Xin and Murata, Tsuyoshi},
  journal={arXiv preprint arXiv:2105.07634},
  year={2021}
}

@article{hamilton2017inductive,
  title={Inductive representation learning on large graphs},
  author={Hamilton, Will and Ying, Zhitao and Leskovec, Jure},
  journal={Advances in neural information processing systems},
  volume={30},
  year={2017}
}

@techreport{arthur2006k,
  title={k-means++: The advantages of careful seeding},
  author={Arthur, David and Vassilvitskii, Sergei},
  year={2006},
  institution={Stanford}
}

@article{yang2023does,
  title={Does graph distillation see like vision dataset counterpart?},
  author={Yang, Beining and Wang, Kai and Sun, Qingyun and Ji, Cheng and Fu, Xingcheng and Tang, Hao and You, Yang and Li, Jianxin},
  journal={Advances in Neural Information Processing Systems},
  volume={36},
  pages={53201--53226},
  year={2023}
}

@inproceedings{xiao2024simple,
  title={Simple graph condensation},
  author={Xiao, Zhenbang and Wang, Yu and Liu, Shunyu and Wang, Huiqiong and Song, Mingli and Zheng, Tongya},
  booktitle={Joint European Conference on Machine Learning and Knowledge Discovery in Databases},
  pages={53--71},
  year={2024},
  organization={Springer}
}

@article{gao2024rethinking,
  title={Rethinking and Accelerating Graph Condensation: A Training-Free Approach with Class Partition},
  author={Gao, Xinyi and Chen, Tong and Zhang, Wentao and Yu, Junliang and Ye, Guanhua and Nguyen, Quoc Viet Hung and Yin, Hongzhi},
  journal={arXiv preprint arXiv:2405.13707},
  year={2024}
}

@article{sfgc,
  title={Structure-free graph condensation: From large-scale graphs to condensed graph-free data},
  author={Zheng, Xin and Zhang, Miao and Chen, Chunyang and Nguyen, Quoc Viet Hung and Zhu, Xingquan and Pan, Shirui},
  journal={Advances in Neural Information Processing Systems},
  volume={36},
  year={2024}
}

@article{kipf2016semi,
  title={Semi-supervised classification with graph convolutional networks},
  author={Kipf, Thomas N and Welling, Max},
  journal={arXiv preprint arXiv:1609.02907},
  year={2016}
}

@inproceedings{opengc,
  title={Graph condensation for open-world graph learning},
  author={Gao, Xinyi and Chen, Tong and Zhang, Wentao and Li, Yayong and Sun, Xiangguo and Yin, Hongzhi},
  booktitle={Proceedings of the 30th ACM SIGKDD Conference on Knowledge Discovery and Data Mining},
  pages={851--862},
  year={2024}
}

@inproceedings{namata2012query,
  title={Query-driven active surveying for collective classification},
  author={Namata, Galileo and London, Ben and Getoor, Lise and Huang, Bert and Edu, U},
  booktitle={10th international workshop on mining and learning with graphs},
  volume={8},
  pages={1},
  year={2012}
}

@article{hu2020open,
  title={Open graph benchmark: Datasets for machine learning on graphs},
  author={Hu, Weihua and Fey, Matthias and Zitnik, Marinka and Dong, Yuxiao and Ren, Hongyu and Liu, Bowen and Catasta, Michele and Leskovec, Jure},
  journal={Advances in neural information processing systems},
  volume={33},
  pages={22118--22133},
  year={2020}
}

@article{zeng2019graphsaint,
  title={Graphsaint: Graph sampling based inductive learning method},
  author={Zeng, Hanqing and Zhou, Hongkuan and Srivastava, Ajitesh and Kannan, Rajgopal and Prasanna, Viktor},
  journal={arXiv preprint arXiv:1907.04931},
  year={2019}
}

@article{bronstein2017geometric,
  title={Geometric deep learning: going beyond euclidean data},
  author={Bronstein, Michael M and Bruna, Joan and LeCun, Yann and Szlam, Arthur and Vandergheynst, Pierre},
  journal={IEEE Signal Processing Magazine},
  volume={34},
  number={4},
  pages={18--42},
  year={2017},
  publisher={IEEE}
}

@article{luo2024classic,
  title={Classic gnns are strong baselines: Reassessing gnns for node classification},
  author={Luo, Yuankai and Shi, Lei and Wu, Xiao-Ming},
  journal={Advances in Neural Information Processing Systems},
  volume={37},
  pages={97650--97669},
  year={2024}
}

@inproceedings{gao2024graph,
  title={Graph condensation for inductive node representation learning},
  author={Gao, Xinyi and Chen, Tong and Zang, Yilong and Zhang, Wentao and Nguyen, Quoc Viet Hung and Zheng, Kai and Yin, Hongzhi},
  booktitle={2024 IEEE 40th International Conference on Data Engineering (ICDE)},
  pages={3056--3069},
  year={2024},
  organization={IEEE}
}

@inproceedings{liu2023cat,
  title={Cat: Balanced continual graph learning with graph condensation},
  author={Liu, Yilun and Qiu, Ruihong and Huang, Zi},
  booktitle={2023 IEEE International Conference on Data Mining (ICDM)},
  pages={1157--1162},
  year={2023},
  organization={IEEE}
}

\clearpage
\appendix
\section{Proof of Theorems}
\label{app:propositions}

\subsection{Proof of Theorems~\ref{theo:training_stage}}
\label{app:proof_training_stage}
\begin{theorem-nonumber}
    The prediction distance in the training stage is bounded by the sum of representation distance and parameter distance.
    \begin{equation}
    \| \mathcal{K}(\hat{\mathbf{Y}}) - \hat{\mathbf{Y}}' \| 
    \leq  \| \mathcal{K}(\mathbf{F}) - \mathbf{F}' \| \cdot \| \mathbf{W}' \| + \| \mathbf{F} \| \cdot \| \mathbf{W} - \mathbf{W}' \| 
    \end{equation}
    where $\|\cdot\|$ denotes the L2 norm and $\mathcal{K}(\cdot)$ can be any projection function that aligns the dimensions of $\hat{{\bf Y}}$ and $\hat{{\bf Y}}'$ or ${\bf F}$ and ${\bf F}'$.
\end{theorem-nonumber}

\begin{proof}
To preserve the training data information to maintain the performance of GNNs, we focus on matching the model predictions on the original graph $G$ and its condensed counterpart $G'$. Since $\mathcal{K}(\cdot)$ only aligns the first dimensions, we have $\mathcal{K}(\hat{\mathbf{Y}}) = \mathcal{K}(\mathbf{F})\mathbf{W}$. Therefore, the expression becomes:
\begin{equation}
    \begin{aligned}
    &\| \mathcal{K}(\hat{\mathbf{Y}}) - \hat{\mathbf{Y}}' \| \\
    = &\| \mathcal{K}(\mathbf{F})\mathbf{W} - \mathbf{F}'\mathbf{W}' \| \\
    = & \| \mathcal{K}(\mathbf{F}) \mathbf{W} - \mathcal{K}(\mathbf{F}) \mathbf{W}' + \mathcal{K}(\mathbf{F}) \mathbf{W}' - \mathbf{F}’ \mathbf{W}' \| \\
    \leq &\| \mathcal{K}(\mathbf{F}) (\mathbf{W} - \mathbf{W}') \| + \| (\mathcal{K}(\mathbf{F}) -  \mathbf{F}')\mathbf{W}' \| \\
    \leq & \| \mathcal{K}(\mathbf{F}) \| \cdot \| \mathbf{W} - \mathbf{W}' \| + \| \mathcal{K}(\mathbf{F}) - \mathbf{F}' \| \cdot \| \mathbf{W}' \|
\end{aligned}
\end{equation}
The objective in training stage is to minimizing $\| \mathcal{K}(\hat{\mathbf{Y}}) - \hat{\mathbf{Y}}' \|$, which can be formulated as:
\begin{equation}
    \begin{aligned}
    &\arg \min_{\hat{\mathbf{Y}}'} \| \mathcal{K}(\hat{\mathbf{Y}}) - \hat{\mathbf{Y}}' \|\\
    =&\arg \min_{\hat{\mathbf{Y}}'} \| \mathcal{K}(\mathbf{F}) \| \cdot \| \mathbf{W} - \mathbf{W}' \| + \| \mathcal{K}(\mathbf{F}) - \mathbf{F}' \| \cdot \| \mathbf{W}' \|\\
    \end{aligned}
\end{equation}
Note that $\|\mathbf{F}\|$ is a constant, and the weight matrix $\| \mathbf{W}' \|$ is naturally constrained due to regularization techniques during model optimization to control its magnitude. Then, we have:
\begin{equation}
    \begin{aligned}
    &\arg \min_{\hat{\mathbf{Y}}'} \| \mathcal{K}(\hat{\mathbf{Y}}) - \hat{\mathbf{Y}}' \|\\
    \approx&\arg \min_{\mathbf{F}'}  \underbrace{\| \mathbf{W} - \mathbf{W}' \|}_{\text{Parameter Distance}} + \underbrace{\| \mathcal{K}(\mathbf{F}) - \mathbf{F}' \|}_{\text{Representation Distance}}
    \end{aligned}
\end{equation}

Therefore, Theorem~\ref{theo:training_stage} indicates that by minimizing the \textbf{representation and parameter distances}, the predictions derived from the condensed graph can be close to those of the original graph.

This completes the proof. 
\end{proof}

\subsection{Proof of Theorems~\ref{theo:test_stage}}
\label{app:proof_test_stage}
\begin{theorem-nonumber}
    The test prediction error of the GNN trained of $G'$ is bounded by the test prediction error of the GNN trained on $G$ plus the parameter distance, as formularized by 
    \begin{equation*}
        \| \mathbf{Y} - \hat{\mathbf{Y}}'' \| \leq \| \mathbf{Y} - \mathbf{F}\mathbf{W} \| + \|\mathbf{F}\| \cdot \| \mathbf{W}- \mathbf{W}'\|
    \end{equation*}
\end{theorem-nonumber}

\begin{proof}
At the test stage, our condensation goal is to ensure that the GNN model, trained on condensed training data $G'$, generalizes effectively to the test graph, i.e., achieving a low prediction error on the test data. 
\begin{equation}
\begin{aligned}
    &\| \mathbf{Y} - \hat{\mathbf{Y}}'' \|  \\
    = &\| \mathbf{Y} - \mathbf{F} \mathbf{W}' \|\\
    = & \| \mathbf{Y} - \mathbf{F} \mathbf{W} + \mathbf{F} \mathbf{W} - \mathbf{F} \mathbf{W}' \| \\
    \leq &\| \mathbf{Y} - \mathbf{F} \mathbf{W} \| + \| \mathbf{F} (\mathbf{W} - \mathbf{W}') \| \\
    \leq &\| \mathbf{Y} - \mathbf{F} \mathbf{W} \| + \| \mathbf{F}\| \cdot \|\mathbf{W} - \mathbf{W}' \|
\end{aligned}
\end{equation}
This inequality incorporates both the original test prediction error and the parameter distance. The objective in testing stage is to minimizing $\| \mathbf{Y} - \hat{\mathbf{Y}}''\|$, which can be formulated as:
\begin{equation}
    \begin{aligned}
    &\arg \min_{\hat{\mathbf{Y}}''} \| \mathbf{Y} - \hat{\mathbf{Y}}'' \|\\
    \approx&\arg \min_{\hat{\mathbf{Y}}'} \| \mathbf{Y} - \mathbf{F} \mathbf{W} \| + \| \mathbf{F}\| \cdot \|\mathbf{W} - \mathbf{W}' \|\\
    \approx&\arg \min_{\hat{\mathbf{Y}}'} \underbrace{\|\mathbf{W} - \mathbf{W}'\|}_{\text{Parameter Distance}} \\
    \end{aligned}
\end{equation}
It indicates that by reducing the \textbf{parameter distance} $\| \mathbf{W}- \mathbf{W}'\|$, the test prediction error becomes more tightly bounded, assuming that the original test prediction error \( \| \mathbf{Y} - \mathbf{F}\mathbf{W}\| \) and the propagated feature matrix \( \mathbf{F} \) remain constant. 

This completes the proof. 
\end{proof}

\subsection{Proof of Theorem \ref{theo:bound_of_parameter}}
\label{app:proof_bound_of_parameter}
\begin{theorem-nonumber}
    The parameter distance can be bounded by the following inequality:
    \begin{equation}
        \| \mathbf{W} - \mathbf{W}' \| \leq \mathcal{C}(\max (\text{diag}(\mathbf{\mathbf{P}^\top\mathbf{P}})))^2,
    \end{equation}
    where $\mathcal{C}=\frac{\|\mathbf{F} \| \cdot \| \mathbf{Y}\|({\lambda_{\min}(\mathbf{F}^\top\mathbf{F})}+\|\mathbf{F} \|)}{{\lambda^2_{\min}(\mathbf{F}^\top\mathbf{F})}}$ is a constant, $\mathbf{P}^\top\mathbf{P} \in \mathbb{R}^{N' \times N'}$ is a diagonal matrix with each diagonal entry corresponding to how many original nodes are assigned to each synthetic node. 
\end{theorem-nonumber}

\begin{proof}
We aim to establish that clustering effectively bounds the error introduced by parameter matching and representation difference. The proofproceeds as follows:
\paragraph{Bounding the Parameter Matching Error \( \| \mathbf{W} - \mathbf{W}' \| \):}
Consider the weight matrices for the SGC and clustering-based methods
\[
\mathbf{W} = (\mathbf{F}^\top \mathbf{F})^{-1} \mathbf{F}^\top \mathbf{Y}, \quad \mathbf{W}' = (\mathbf{C}^\top \mathbf{C})^{-1} \mathbf{C}^\top \mathbf{Y}'
\]
where
\[
\mathbf{C} = (\mathbf{P}^\top\mathbf{P})^{-1} \mathbf{P}^\top \mathbf{F}, \quad \mathbf{Y}' = (\mathbf{P}^\top\mathbf{P}))^{-1} \mathbf{P}^\top \mathbf{Y}
\]

The difference between \( \mathbf{W} \) and \( \mathbf{W}' \) is
\[
\| \mathbf{W} - \mathbf{W}' \| = \left\| (\mathbf{F}^\top \mathbf{F})^{-1} \mathbf{F}^\top \mathbf{Y} - (\mathbf{C}^\top \mathbf{C})^{-1} \mathbf{C}^\top \mathbf{Y}' \right\|
\]

By substituting $\mathbf{C}$ and $\mathbf{Y}'$, we can express \( \mathbf{W}' \) in terms of \( \mathbf{F} \) and \( \mathbf{Y} \)
\[
\mathbf{W}' = \left( \mathbf{F}^\top \mathbf{P} (\mathbf{P}^\top\mathbf{P})^{-2} \mathbf{P}^\top \mathbf{F} \right)^{-1} \mathbf{F}^\top \mathbf{P} (\mathbf{P}^\top\mathbf{P})^{-2} \mathbf{P}^\top \mathbf{Y}
\]

Let \( \mathbf{A} = \mathbf{F}^\top \mathbf{F} \) and \( \mathbf{B} = \mathbf{F}^\top \mathbf{P} (\mathbf{P}^\top\mathbf{P})^{-2} \mathbf{P}^\top \mathbf{F} \), then:
\[
\mathbf{W} = \mathbf{A}^{-1} \mathbf{F}^\top \mathbf{Y}, \quad \mathbf{W}' = \mathbf{B}^{-1} \mathbf{F}^\top \mathbf{P} (\mathbf{P}^\top\mathbf{P})^{-2} \mathbf{P}^\top \mathbf{Y}
\]

The difference becomes
\[
\mathbf{W} - \mathbf{W}' = \mathbf{A}^{-1} \mathbf{F}^\top \mathbf{Y} - \mathbf{B}^{-1} \mathbf{F}^\top \mathbf{P} (\mathbf{P}^\top\mathbf{P})^{-2} \mathbf{P}^\top \mathbf{Y}
\]

Similar to the above two proofs, we add and subtract a term \( \mathbf{B}^{-1} \mathbf{F}^\top \mathbf{Y} \) and rewrite the difference by
\begin{align*}
        \mathbf{W} - \mathbf{W}' = &\left( \mathbf{A}^{-1} - \mathbf{B}^{-1} \right) \mathbf{F}^\top \mathbf{Y} \\
&+ \mathbf{B}^{-1} \mathbf{F}^\top \left( \mathbf{I} - \mathbf{P} (\mathbf{P}^\top\mathbf{P})^{-2} \mathbf{P}^\top \right) \mathbf{Y}
\end{align*}

Considering the norms, we have
    \begin{align*}
\| \mathbf{W} - \mathbf{W}' \| \leq &\| \mathbf{A}^{-1} - \mathbf{B}^{-1} \| \cdot \| \mathbf{F}^\top \mathbf{Y} \| \\
&+ \| \mathbf{B}^{-1} \| \cdot \| \mathbf{F}^\top (\mathbf{I} - \mathbf{P} (\mathbf{P}^\top\mathbf{P})^{-2} \mathbf{P}^\top) \mathbf{Y} \|
\end{align*}

We will now bound each term independently.

\textbf{Bounding the First Term}
\begin{align*}
        \| \mathbf{A}^{-1} - \mathbf{B}^{-1} \| \cdot \| \mathbf{F}^\top \mathbf{Y} \|
\end{align*}

Based on the norms of matrix inequality, we have
\[
\| \mathbf{A}^{-1} - \mathbf{B}^{-1} \| \leq \| \mathbf{A}^{-1} \| \cdot \| \mathbf{B} - \mathbf{A} \| \cdot \| \mathbf{B}^{-1} \|
\]
Then, according to
\[
\| \mathbf{A}^{-1} \| = \frac{1}{\lambda_{\min}(\mathbf{A})}, \quad \| \mathbf{B}^{-1} \| = \frac{1}{\lambda_{\min}(\mathbf{B})}
\]
and assuming \( \lambda_{\min}(\mathbf{B}) \geq \frac{1}{(\max_k (\mathbf{P}^\top\mathbf{P})_{kk})^2} \lambda_{\min}(\mathbf{A}) \), we have
\[
\| \mathbf{A}^{-1} \| \cdot \| \mathbf{B}^{-1} \| \leq \frac{(\max_k (\mathbf{P}^\top\mathbf{P})_{kk})^2}{\lambda_{\min}(\mathbf{A})^2}
\]

Bounding \( \| \mathbf{B} - \mathbf{A} \| \):
\begin{align*}
    \mathbf{B} - \mathbf{A} &= \mathbf{F}^\top \mathbf{P} (\mathbf{P}^\top\mathbf{P})^{-2} \mathbf{P}^\top \mathbf{F} - \mathbf{F}^\top \mathbf{F} \\
    &= -\mathbf{F}^\top (\mathbf{I} - \mathbf{P} (\mathbf{P}^\top\mathbf{P})^{-2} \mathbf{P}^\top) \mathbf{F}
\end{align*}
Taking norms:
\begin{align*}
\| \mathbf{B} - \mathbf{A} \| &= \| \mathbf{F}^\top (\mathbf{P} (\mathbf{P}^\top\mathbf{P})^{-2} \mathbf{P}^\top - \mathbf{I}) \mathbf{F} \| \\
&\leq \| \mathbf{F} \|^2 \cdot \| \mathbf{I} - \mathbf{P} (\mathbf{P}^\top\mathbf{P})^{-2} \mathbf{P}^\top \|
\end{align*}
Since \( \| \mathbf{I} - \mathbf{P} (\mathbf{P}^\top\mathbf{P})^{-2} \mathbf{P}^\top \| \leq 1 \), the inequality can be further simplified to
\(\| \mathbf{B} - \mathbf{A} \| \leq \| \mathbf{F} \|^2\).

Combining the above:
\[
\| \mathbf{A}^{-1} - \mathbf{B}^{-1} \| \cdot \| \mathbf{F}^\top \mathbf{Y} \| \leq \frac{(\max_k (\mathbf{P}^\top\mathbf{P})_{kk})^2}{\lambda_{\min}(\mathbf{A})^2} \cdot \| \mathbf{F} \|^2 \cdot \| \mathbf{Y} \|
\]

\textbf{Bounding the Second Term}
\begin{align*}
    \| \mathbf{B}^{-1} \| \cdot \| \mathbf{F}^\top (\mathbf{I} - \mathbf{P} (\mathbf{P}^\top\mathbf{P})^{-2} \mathbf{P}^\top) \mathbf{Y} \|
\end{align*}

Based on the proof above, we first have
\[
\| \mathbf{B}^{-1} \| = \frac{1}{\lambda_{\min}(\mathbf{B})} \leq \frac{(\max_k (\mathbf{P}^\top\mathbf{P})_{kk})^2}{\lambda_{\min}(\mathbf{A})}
\]

 Since \( \| \mathbf{I} - \mathbf{P} (\mathbf{P}^\top\mathbf{P})^{-2} \mathbf{P}^\top \| \leq 1 \), 
\[
\| \mathbf{F}^\top (\mathbf{I} - \mathbf{P} (\mathbf{P}^\top\mathbf{P})^{-2} \mathbf{P}^\top) \mathbf{Y} \| \leq \| \mathbf{F} \| \cdot \| \mathbf{Y} \|
\]

Combining these bounds:
\begin{align*}
        &\| \mathbf{B}^{-1} \| \cdot \| \mathbf{F}^\top (\mathbf{I} - \mathbf{P} (\mathbf{P}^\top\mathbf{P})^{-2} \mathbf{P}^\top) \mathbf{Y} \| \\
        &\leq \frac{(\max_k (\mathbf{P}^\top\mathbf{P})_{kk})^2}{\lambda_{\min}(\mathbf{A})} \cdot \| \mathbf{F} \| \cdot \| \mathbf{Y} \|
\end{align*}

\paragraph{Combining Both Terms:}
Adding the bounds for both terms, we obtain:
\begin{align*}
\| \mathbf{W} - \mathbf{W}' \| \leq & \frac{(\max_k (\mathbf{P}^\top\mathbf{P})_{kk})^2}{\lambda_{\min}(\mathbf{A})^2} \cdot \| \mathbf{F} \|^2 \cdot \| \mathbf{Y} \| \\
&\quad + \frac{(\max_k (\mathbf{P}^\top\mathbf{P})_{kk})^2}{\lambda_{\min}(\mathbf{A})} \cdot \| \mathbf{F} \| \cdot \| \mathbf{Y} \| \\
=&\max_k (\mathbf{P}^\top\mathbf{P})_{kk}) \cdot (\frac{\| \mathbf{F} \|^2 \cdot \| \mathbf{Y} \|}{\lambda_{\min}(\mathbf{A})^2} + \frac{\| \mathbf{F} \| \cdot \| \mathbf{Y} \|}{\lambda_{\min}(\mathbf{A})}) \\
=&\mathcal{C}(\max (\text{diag}(\mathbf{\mathbf{P}^\top\mathbf{P}})))^2
\end{align*}
where $\mathcal{C}=\frac{\|\mathbf{F} \| \cdot \| \mathbf{Y}\|({\lambda_{\min}(\mathbf{F}^\top\mathbf{F})}+\|\mathbf{F} \|)}{{\lambda^2_{\min}(\mathbf{F}^\top\mathbf{F})}}$ is a constant, $\mathbf{P}^\top\mathbf{P} \in \mathbb{R}^{N' \times N'}$ is a diagonal matrix with each diagonal entry corresponding to how many original nodes are assigned to each synthetic node. 

Our objective is to minimize the parameter distance, which can be reformulated by:
\begin{equation}
    \begin{aligned}
    &\arg \min_{\mathbf{W}'} \| \mathbf{W} - \mathbf{W}' \|\\
    \approx&\arg \min_{\mathbf{P}} \mathcal{C}(\max (\text{diag}(\mathbf{\mathbf{P}^\top\mathbf{P}})))^2\\
    \end{aligned}
\end{equation}

This completes the proof.   
\end{proof}

\section{Method Pipeline}
\begin{figure}[h]
    \centering
    \includegraphics[width=0.9\linewidth]{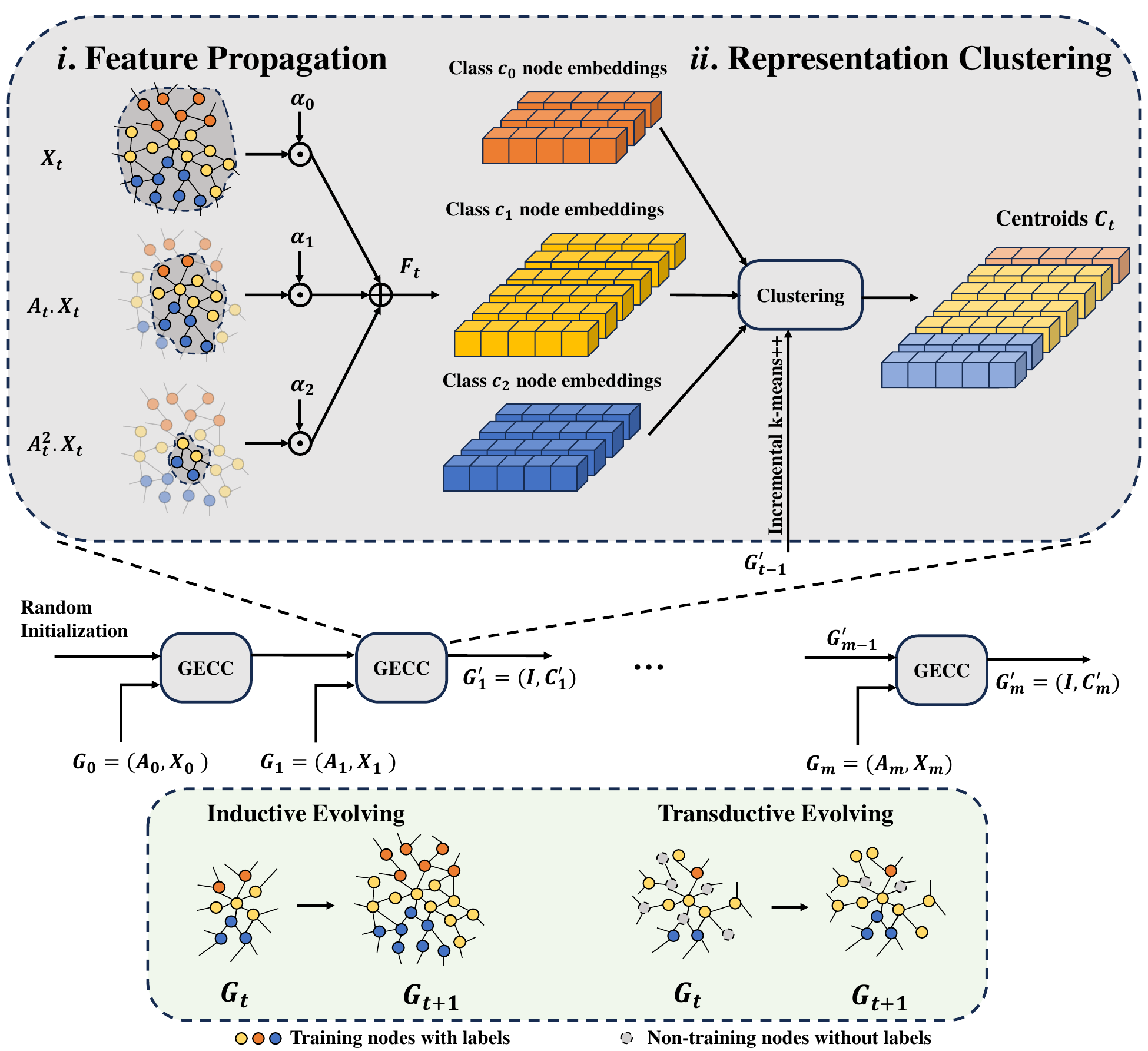}
    \caption{Illustration of the GECC framework. The lower part of the figure depicts the two evolving settings: in the \textbf{inductive} setting, the graph structure evolves over time, while in the \textbf{transductive} setting, only the training nodes (labels) change, with the graph structure and non-training nodes remaining unchanged.}

    \label{fig:main}

\end{figure}

The proposed GECC framework is illustrated in Figure~\ref{fig:main}. GECC takes the current graph \( G_t = (\mathbf{X}_t, \mathbf{A}_t) \) along with the centroids from the previous time step to perform condensation and generate the condensed graph \( G'_t = (\mathbf{I}, \mathbf{C}'_t) \). The lower part of the figure also depicts the two evolving settings: in the \textbf{inductive} setting, the graph structure evolves over time, while in the \textbf{transductive} setting, only the training nodes (labels) change, whereas the graph structure and non-training nodes remain unchanged.
\section{Symbol Table}
All the symbols used in this paper have been summarized in Table~\ref{tab:symbol}
\begin{table*}[ht]
\centering
\caption{Notations}
\label{tab:symbol}
\begin{tabular}{@{}ll@{}}
\toprule
\textbf{Symbol} & \textbf{Description} \\
\midrule
\multicolumn{2}{l}{\textit{General Graph Notations}} \\
$G, G_t$ & An static graph and a snapshot of an evolving graph at time step $t$. \\
$G', G'_t$ & The condensed graph from the static graph and the evolving graph at time step $t$. \\
$\mathcal{V}, \mathcal{E}$ & The set of nodes and edges in a graph. \\
$N, E$ & The total number of nodes and edges. \\
$\mathbf{X}, \mathbf{X}_t$ & The node feature matrix of the static graph and evolving graph at time $t$. \\
$\mathbf{X}', \mathbf{X}'_t$ & The feature matrix of the condensed graph (i.e., the centroids). \\
$\mathbf{A}, \mathbf{A}_t$ & The adjacency matrix of the original graph and at time $t$. \\
$\mathbf{A}',\mathbf{A}'_t$ & The adjacency matrix of the condensed graph (set to $\mathbf{I}$). \\
$\mathbf{y}, \mathbf{Y}, \mathbf{Y}_t$ & The numeric and one-hot encoded label vectors/matrices of the original graphs. \\
$\mathbf{Y}', \mathbf{Y}_t'$ & The one-hot encoded label of the condensed nodes. \\
$c$ & The total number of classes. \\
$\mathbf{L}, \mathbf{D}$ & The graph Laplacian and degree matrices. \\
$\mathcal{F}$ & The update function to generate a new condensed graph. \\
\midrule
\multicolumn{2}{l}{\textit{Evolving Graph Condensation}} \\
$B_t$ & A batch of new graph data arriving at time step $t$. \\
$m$ & The total number of time steps in the evolving sequence. \\

$r$ & The reduction rate, determining the size of the condensed graph. \\
\midrule
\multicolumn{2}{l}{\textit{GECC Algorithm Notations}} \\
$\tilde{A}_t, \hat{A}_t$ & The adjacency matrix with self-loops and its normalized version. \\
$K$ & The maximum depth for feature propagation. \\
$F_t$ & The propagated node feature matrix at time $t$. \\
$\alpha_k$ & Learnable weight for combining features from propagation depth $k$. \\
$n_{t,k}$ & The number of nodes belonging to class $k$ at time $t$. \\
$M, M_k$ & The total number of condensed nodes (clusters) and per class. \\
$P_t$ & The cluster assignment matrix at time $t$. \\
$C_t, C_{t-1}$ & The set of cluster centroids at time $t$ and $t-1$. \\
$J(\cdot, \cdot)$ & The balanced Sum of Squared Errors (SSE) objective function. \\
$D(x_i, C)$ & The minimum Euclidean distance from a node $x_i$ to a set of centroids $C$. \\
\bottomrule
\end{tabular}
\end{table*}
\clearpage
\begin{algorithm*}
\caption{GECC Node Condensation}
\begin{algorithmic}[1]
\Require Adjacency matrix $A_t$, feature matrix $X_t$, label vector $y_t$, reduction rate $r$, label matrix $Y'_t \gets$ predefined using class distribution in $Y_t$
\Ensure Condensed node representations $X'_t$

\State \textbf{// Feature Propagation (Eq. 5)}
\State $\tilde{A}_t \gets A_t + I$ \Comment{Add self-loops}
\State $\tilde{D}_t \gets \text{diag}(\sum_j \tilde{A}_{t,i,j})$
\State $\hat{A}_t \gets \tilde{D}_t^{-\frac{1}{2}} \tilde{A}_t \tilde{D}_t^{-\frac{1}{2}}$
\For{$k = 0$ to $K$}
    \State $F_{t,k} \gets \hat{A}_t^k X_t$
\EndFor
\State $F_t \gets \sum_{k=0}^{K} \alpha_k F_{t,k}$ \Comment{Combine with learnable weights $\alpha_k$}

\State \textbf{// Class-wise Clustering Setup}
\For{each class $k = 1, \ldots, c$}
    \State $n_{t,k} \gets$ number of labeled nodes in class $k$
    \State $M_k \gets \lfloor n_{t,k} \times r \rfloor$ \Comment{Eq. before clustering}
\EndFor
\State $M \gets \sum_{k=1}^{c} M_k$

\State \textbf{// Cluster Assignment Matrix $P_t$ and Balanced SSE (Eq. 5)}
\For{each clustering repetition}
    \State Apply $k$-means or fuzzy $c$-means on $F_t$ to get $P_t \in \mathbb{R}^{n_t \times M}$
    \State Compute cluster centroids: $C_t = D_{P_t}^{-1} P_t^\top F_t$ \Comment{Eq. 6}
    \State Compute balanced SSE loss:
    \[
    J(P_t, C_t) = \|F_t - P_t C_t\|^2 + \|P_t^\top \mathbf{1} - u\|^2 \tag{5}
    \]
\EndFor
\State Choose clustering with lowest $J(P_t, C_t)$

\State \textbf{// Probabilistic Centroid Initialization (Diverse Coverage)}
\State Initialize $C_{t-1} = \{\}$ 
\For{each class $c$ and node $x_i \in F_t$}
    \State $D(x_i, C_{t-1}) = \min_{c \in C_{t-1}} \|x_i - c\|_2$
\EndFor
\State Sample new centroid $x_i$ with probability:
\[
P(x_i) = \frac{D(x_i, C_{t-1})^2}{\sum_j D(x_j, C_{t-1})^2}
\]
\State $C_t \gets C_{t-1} \cup \{C_\Delta\}$

\State \textbf{// Final Assignment}
\For{each node $x_i$}
    \State $\text{Cluster}(x_i) = \arg\min_{c \in C_t} \|x_i - c\|_2$
\EndFor

\State $X'_t \gets C_t$

\end{algorithmic}
\end{algorithm*}

\section{Complexity analysis}~\label{app:complexity}
We analyze the complexity of each component in \textsc{GECC} as follows.
\textbf{First}, feature propagation takes \(\mathcal{O}(Ke_td)\) time, where \(K\) is the propagation depth, \(e_t\) is the number of edges at time step \(t\), and \(d\) is the feature dimension.
\textbf{Second}, hard \(k\)-means with \(M\) clusters requires \(\mathcal{O}(n_tUMd)\) per iteration (\(U\) is the number of iterations), whereas soft \(k\)-means also costs \(\mathcal{O}(n_tUMd)\) for distance calculations but plus an additional \(\mathcal{O}(n_tM)\) overhead to normalize fractional memberships. 
\textbf{Third}, the balanced SSE calculation adds only \(\mathcal{O}(n_tM)\) overhead, as the repetitive clustering can be done in parallel. 
\textbf{Finally}, in an evolving setting, incremental \(k\)-means++ only considers newly arrived nodes, reducing initialization cost to \(\mathcal{O}(m|C_{t-1}|d)\), where \(m\) is the number of new nodes and \(|C_{t-1}|\) is the number of existing centroids. 
Gathering all components, the total complexity is dominated by \(\mathcal{O}(n_tUMd)\) from the clustering procedure and \(\mathcal{O}(Ke_td)\) from feature propagation, both of which scale \textbf{linearly} with the number of nodes \(n_t\) and edge count \(e_t\), given that the size of condensed graph $M$ is negligible in terms of order of magnitude compared to $n_t$.
\section{Experimental Details}\label{app:exp}
\begin{table*}[ht]
\centering
\caption{General Dataset Information}
\label{tab:statistics}
\begin{tabular}{lrrrrrrr}
\toprule
\textbf{Dataset} & 
\textbf{\# Total Nodes} & 
\textbf{\# Total Edges} & 
\textbf{\# Training} & 
\textbf{\# Validation} & 
\textbf{\# Test} & 
\textbf{\# Classes} & 
\textbf{Trans./Ind.} \\
\midrule
\textit{Citeseer}      & 3{,}327     & 4{,}732        & 120      & 500       & 1{,}000   & 6  & Transductive \\
\textit{Cora}          & 2{,}708     & 5{,}429        & 140      & 500       & 1{,}000   & 7  & Transductive \\
\textit{Pubmed}        & 19{,}717    & 88{,}648       & 60       & 500       & 1{,}000   & 3  & Transductive \\
\textit{Flickr}        & 89{,}250    & 899{,}756      & 44{,}625 & 22{,}312  & 22{,}313  & 7  & Inductive    \\ \midrule
\textit{Ogbn-arxiv (-real)}    & 169{,}343   & 1{,}166{,}243  & 90{,}941 & 29{,}799  & 48{,}603  & 40 & Transductive \\
\textit{Ogbn-products} & 2{,}449{,}029 & 39{,}561{,}252 & 196{,}615 & 39{,}323 & 2{,}213{,}091 & 47 & Transductive \\
\textit{Reddit}        & 232{,}965   & 57{,}307{,}946 & 153{,}932 & 23{,}699  & 55{,}334  & 41 & Inductive    \\
\bottomrule
\end{tabular}
\end{table*}

\begin{figure}[t]
    \centering
    \includegraphics[width=\linewidth]{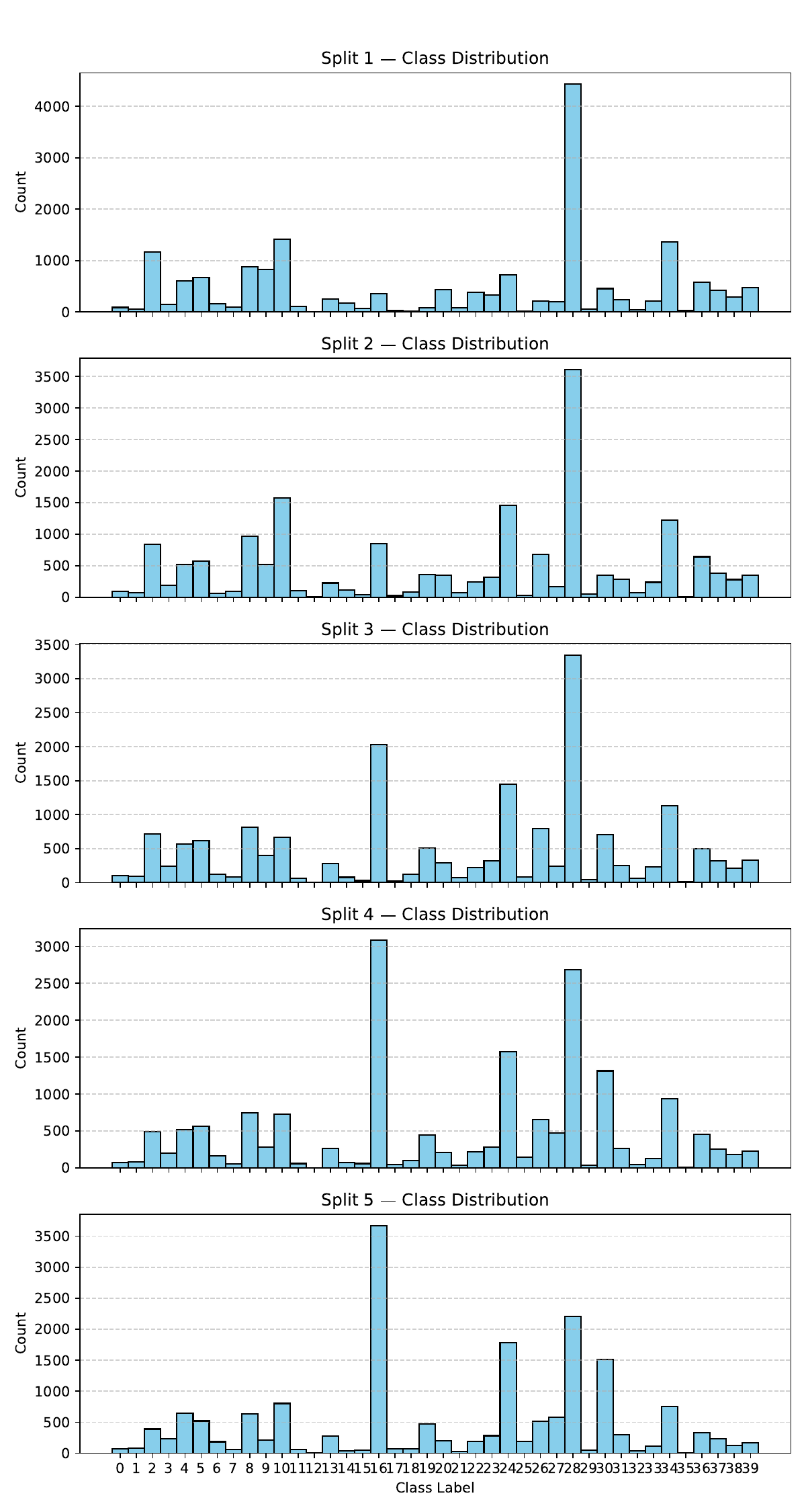}
    \caption{Class distribution across five splits of the \textit{Ogbn-arxiv-real} dataset.} 

    \label{fig:dist}
\end{figure}

\subsection{Dataset Statistics}
\label{app:statistics}
In line with most GC studies, we utilize seven datasets in total: five transductive datasets—Citeseer, Cora \citep{kipf2016semi}, Pubmed \citep{namata2012query}, Ogbn-arxiv, and Ogbn-products \citep{hu2020open}—and two inductive datasets, Flickr and Reddit \citep{zeng2019graphsaint}. Each graph is randomly split, ensuring a consistent class distribution. The details of the datasets statistics are shown in Table \ref{tab:statistics}. We list all evolving information in Table~\ref{tab:split_reduction}, rows above the midline correspond to smaller datasets, and rows below it correspond to larger ones.
The real-world evolving information of \textit{Ogbn-arxiv-real} is shown in Table~\ref{tab:real}. The distribution of the number of nodes in each class is shown in Figure \ref{fig:dist}. 
\begin{table}[t!]
\centering
\caption{Training Set Splits of \textbf{Ogbn-arxiv-real} by Year Range and Node Count}
\label{tab:real}
\begin{tabular}{@{}lcr@{}}
\toprule
\textbf{Split No.} & \textbf{Year Range} & \multicolumn{1}{c}{\textbf{Number of Nodes}} \\
\midrule
1 & 1971 -- 2007 & 4,980 \\
2 & 2008 -- 2010 & 7,994 \\
3 & 2011 -- 2012 & 10,862 \\
4 & 2013 -- 2013 & 8,135 \\
5 & 2014 -- 2017 & 58,970 \\
\bottomrule
\end{tabular}
\end{table}

Reduction rate $r$ is defined as (\#nodes in synthetic set)/(\#nodes in training set) while $r_w$ is (\#nodes in synthetic set)/(\#nodes of whole graph visible in training stage). The whole graph visible in training stage means the full graph dataset for transductive setting but only the training graph for inductive setting. 
\begin{table}[ht!]
\caption{Split and reduction rate information. The ``\# Train Nodes'' and ``\# Syn Nodes'' columns denote the number of newly added training nodes and synthetic nodes at each time step, respectively.}
\label{tab:split_reduction}
\resizebox{0.47\textwidth}{!}{
\begin{tabular}{lrrrr}
\toprule
\textbf{Dataset} & 
\textbf{\# Train Nodes} & 
\textbf{\# Syn Nodes} & 
\textbf{$r$ (Train)} & 
\textbf{$r_w$ (Whole)} \\
\midrule
\textit{Citeseer}      & 24             & 12  & 0.5    & 1.80  \\
\textit{Cora}          & 28             & 14  & 0.5    & 2.60  \\
\textit{Pubmed}        & 12             & 6   & 0.5    & 0.15  \\
\midrule
\textit{Flickr}        & ~8{,}920 & 90  & 0.01   & 1.00  \\
\textit{Ogbn-arxiv (-real)}    & ~18{,}190 & 182 & 0.01   & 0.50  \\
\textit{Ogbn-products} & ~39{,}330 & 394 & 0.01   & 0.08  \\
\textit{Reddit}        & ~30{,}790 & 31  & 0.001  & 0.10  \\
\bottomrule
\end{tabular}}
\end{table}

\subsection{Platform and Hardware Information}
To efficiently execute the clustering algorithm, we run it on Intel(R) Xeon(R) Platinum 8260 CPUs @ 2.40GHz using NumPy~\cite{numpy}, while the downstream GNN evaluations are conducted on a cluster equipped with a mix of Tesla A100 40GB/V100 32GB GPUs for large datasets and K80 12GB GPUs for smaller datasets. All GNN models are implemented using the PyG package~\cite{pyg}.

\subsection{Baselines Selection}
To establish a fair benchmark, we selected recent state-of-the-art GC methods that emphasize both effectiveness and efficiency. Some recent methods, such as MCond, CGC, and GCPA, were excluded due to the unavailability of their code at the time of paper writing. For the selected approaches, we chose the best representatives from each category: GCondX for gradient matching, GCDM and SimGC for distribution matching, and GEOM for trajectory matching. We implemented these methods using the latest GraphSlim package\footnote{\url{https://github.com/Emory-Melody/GraphSlim/tree/main}}, except for SimGC\footnote{\url{https://github.com/BangHonor/SimGC}} and GEOM\footnote{\url{https://github.com/NUS-HPC-AI-Lab/GEOM/tree/main}}, for which we used their original source code. We specifically included SimGC because it is the only model-based GC method that can run on Ogbn-products without requiring any modifications.

\subsection{Implementation Details for Variants of GCondX}
As mentioned in Section~\ref{sec:intro} and illustrated in Table~\ref{tab:preliminary}, adapting GCondX to an evolving setting is challenging. The experimental results indicate that, although there is a slight efficiency improvement, there is a significant performance drop due to inheriting previous condensation results. To highlight the difference in speed, we additionally implement an early stopping criterion with a patience of 3 for intermediate evaluation; that is, if no improvement in validation performance is observed for 3 consecutive evaluations, the condensation process is terminated.
\subsection{Hyperparameters}\label{app:hyper}
Compared to existing work and benchmarks in GC, we perform a moderate hyperparameter search on validation set, as detailed in Section~\ref{sec:hyper}. The final results are presented in Table~\ref{tab:hyper}. During hyperparameter optimization (HPO), we observe that inheriting clustering centroids results in an approximate 1\% absolute performance drop for \textit{Flickr} and \textit{Ogbn-arxiv}. Therefore, we also treat the use of incremental $k$-means++ as a tunable hyperparameter. Additionally, certain datasets do not perform well under a single hyperparameter configuration. To address this, we employ two distinct hyperparameter sets tuned on the first and last time step, and select the better-performing one during the evolving. The two sets of hyperparameters are represented split by "/". We do not tune the hyperparameters for Ogbn-arxiv-real and simply copy it from Ogbn-arxiv. For all baselines, we use the best hyperparameters reported in their respective papers, as implemented in GC4NC~\cite{gong2024gc4nc}.

Table~\ref{tab:hyper} reveals that the optimal hyperparameters offer meaningful insights. \textbf{First}, during the early evolution stage, graphs exhibit higher heterophily compared to later stages. For example, on the \textit{Cora} dataset, $\alpha_1=-0.3$ in the early phase contrasts with $\alpha_1=0.9$ later. This pattern likely arises because, in the early stages of a graph, groups have not yet formed; links appear more randomly, making it challenging for nodes to find similar counterparts.
\textbf{Second}, it is noteworthy that some datasets do not rely on second-hop information. This observation is contrary to previous studies~\cite{wu2019simplifying,luo2024classic} that recommend using at least 2-hop propagation. We conjecture that the representation clustering process itself acts as an additional step of feature propagation.
\textbf{Finally}, weight decay emerges as a critical factor for the performance of downstream models, suggesting that future work should pay closer attention to its optimization.
\begin{table*}[ht!]
\centering
\caption{The test accuracy of GC methods on various datasets.
"Non-Evolving" displays the test accuracy at the final time step (largest possible graph).
"Evolving" shows the average test accuracy over five time-steps.
Each result includes the mean accuracy $\pm$ standard deviation (Std.) from 10 runs. The "Whole" column refers to the results obtained by running standard GCN training and testing. "OOM" indicates an Out-of-Memory error during the computation. The best results are marked in \textbf{bold}. The runner-up results are \underline{underlined}.}
\resizebox{\textwidth}{!}{%
\begin{tabular}{lc|ccc|cccccc|c}
\toprule
\textbf{Dataset} & \textbf{Setting} 
  & \textbf{Random} 
  & \textbf{Herding} 
  & \textbf{Kcenter} 
  & \textbf{GCondX} 
  & \textbf{GCond} 
  & \textbf{GCDM} 
  & \textbf{SimGC}
  & \textbf{GEOM} 
  & \textbf{GECC} 
  & \multicolumn{1}{c}{\textbf{Whole}} \\
  \midrule
\multirow{2}{*}{CiteSeer} 
  &Non-Evolving& 62.62$\pm$0.63 & 66.66$\pm$0.54 & 59.04$\pm$0.90 & 68.38$\pm$0.45 & 69.35$\pm$0.82 & 72.08$\pm$0.19 & 66.40$\pm$0.15 &  \textcolor{red}{\underline{73.03$\pm$0.31}} & \textcolor{red}{\textbf{73.25$\pm$0.15}} & 72.11 \\
  &Evolving& 50.65$\pm$1.55 & 53.47$\pm$0.98 & 47.99$\pm$1.81 & 50.85$\pm$3.00 & 60.51$\pm$0.86 & \textcolor{blue}{\underline{61.51$\pm$0.53}}& 57.42$\pm$0.21 & 58.95$\pm$0.67 & \textcolor{blue}{\textbf{65.48$\pm$0.76}} & 63.57 \\
\midrule
\multirow{2}{*}{Cora} 
  &Non-Evolving& 72.24$\pm$0.59 & 73.77$\pm$0.93 & 70.55$\pm$1.35 & 78.60$\pm$0.31 & 80.54$\pm$0.67 & 80.68$\pm$0.27 & 79.60$\pm$0.11 & \textcolor{red}{\underline{82.82$\pm$0.17}} & \textcolor{red}{\textbf{82.99$\pm$0.27}} & 81.23 \\
  &Evolving& 58.00$\pm$1.48 & 63.07$\pm$1.43 & 59.90$\pm$1.41 & 67.18$\pm$1.73 & \textcolor{blue}{\underline{77.14$\pm$0.55}} & 74.54$\pm$0.59 & 64.42$\pm$0.19 & 72.56$\pm$0.88 & \textcolor{blue}{\textbf{77.36$\pm$0.41}} & 76.34 \\
\midrule
\multirow{2}{*}{Pubmed} 
  &Non-Evolving& 71.84$\pm$0.66 & 75.53$\pm$0.44 & 74.00$\pm$0.19 & 71.97$\pm$0.53 & 76.46$\pm$0.48 & 77.48$\pm$0.46 & 76.80$\pm$0.23 & \textcolor{red}{\underline{78.49$\pm$0.24}} & \textcolor{red}{\textbf{80.24$\pm$0.27}} & 78.65 \\
  &Evolving& 66.37$\pm$1.25 & 66.31$\pm$1.34 & 64.38$\pm$1.25 & 62.65$\pm$1.20 & 74.26$\pm$0.84 & \textcolor{blue}{\underline{74.49$\pm$0.56}} & 71.38$\pm$0.21 & 70.25$\pm$0.78 &  \textcolor{blue}{\textbf{76.74$\pm$0.27}} & 76.18 \\
\midrule
\multirow{2}{*}{Flickr} 
  &Non-Evolving& 44.68$\pm$0.55 & 45.12$\pm$0.39 & 43.53$\pm$0.59 & 46.58$\pm$0.14 & \textcolor{red}{\textbf{46.99$\pm$0.12}} & 45.88$\pm$0.10 & 41.01$\pm$0.23 & 46.13$\pm$0.22 & \textcolor{red}{\underline{46.63$\pm$0.23}} & 47.53 \\
  &Evolving& 44.70$\pm$0.46 & 44.66$\pm$0.43 & 44.33$\pm$0.49 & \textcolor{blue}{\underline{45.63$\pm$0.78}}& 45.52$\pm$0.49 & 44.98$\pm$0.34 & 41.94$\pm$0.22 & 45.43$\pm$0.39 &  \textcolor{blue}{\textbf{45.78$\pm$0.38}} & 46.97 \\
  \midrule
  \multirow{2}{*}{Ogbn-arxiv} 
  &Non-Evolving& 60.19$\pm$0.52 & 57.70$\pm$0.24 & 58.66$\pm$0.36 & 59.93$\pm$0.54 & 64.23$\pm$0.16 & 60.71$\pm$0.68 & 65.26$\pm$0.26 &  \textcolor{red}{\textbf{69.59$\pm$0.24}} & \textcolor{red}{\underline{66.71$\pm$0.10}} & 69.01 \\
  &Evolving& 56.04$\pm$0.67 & 57.57$\pm$0.48 & 56.21$\pm$0.73 & 60.73$\pm$0.53 & 62.50$\pm$0.36 & 59.98$\pm$0.48 & 64.97$\pm$0.20 & \textcolor{blue}{\textbf{66.30$\pm$0.39}} & \textcolor{blue}{\underline{65.42$\pm$0.14}} & 70.40 \\
\midrule
\multirow{2}{*}{Ogbn-products} 
  &Non-Evolving& 60.19$\pm$0.52 & 57.70$\pm$0.24 & 58.66$\pm$0.36 & OOM & OOM & OOM & \textcolor{red}{\underline{61.71$\pm$0.25}} & OOM & \textcolor{red}{\textbf{66.32$\pm$0.23}} & 73.40 \\
  &Evolving& 41.36$\pm$0.48 & 44.26$\pm$0.61 & 38.93$\pm$0.82 & OOM & OOM & OOM & \textcolor{blue}{\underline{61.93$\pm$0.20}} & OOM & \textcolor{blue}{\textbf{64.03$\pm$0.30}} & 73.88 \\
\midrule
\multirow{2}{*}{Reddit} 
  &Non-Evolving& 55.73$\pm$0.50 & 59.34$\pm$0.70 & 48.28$\pm$0.73 & 88.25$\pm$0.30 & 89.82$\pm$0.10 & 89.96$\pm$0.05 & 90.78$\pm$0.25 & \textcolor{red}{\underline{91.33$\pm$0.13}} & \textcolor{red}{\textbf{91.37$\pm$0.04}} & 93.70 \\
  &Evolving& 51.31$\pm$0.90 & 48.94$\pm$0.70 & 48.53$\pm$1.37 & 79.02$\pm$0.73 & 87.93$\pm$0.22 & 82.68$\pm$0.21 &\textcolor{blue}{\underline{89.85$\pm$0.25}} & 67.91$\pm$0.57 & \textcolor{blue}{\textbf{90.02$\pm$0.07}} & 93.92 \\
\bottomrule
\end{tabular}}
\label{tab:main_app}
\end{table*}
\begin{table*}[ht!]
\centering
\caption{\textbf{Average Runtime (seconds) Across Evolving Times.} The reported reduction time is rigorously computed by excluding the overhead of the data loading and evaluation processes.}
\label{tab:time}
\resizebox{\textwidth}{!}{
\begin{tabular}{lccc|cccccc|c}
\toprule
\textbf{Dataset} & \textbf{Random} & \textbf{Herding} & \textbf{KCenter} & \textbf{GCondX}& \textbf{GCond}& \textbf{GCDM} & \textbf{SimGC}& \textbf{GEOM}& \textbf{GECC} & \textbf{Whole}\\
\midrule
\textit{Citeseer}
  & 0.04
  & 5.73
  & 5.84
  & 505.62
& 654.32
& 217.99
  & 1,680.02
& 1,362.40
& 1.65
& 3.98 \\

\textit{Cora}
  & 0.01
  & 4.20
  & 4.80
  & 331.53
& 1,190.65
& 142.82
  & 1,643.76
& 1,331.43
& 1.72
& 2.10 \\

\textit{Pubmed}
  & 0.02
  & 9.00
  & 7.18
  & 246.68
& 502.12
& 311.37
  & 1,654.23
& 995.21
& 1.42
& 5.76 \\ \midrule

\textit{Flickr}
  & 0.02
  & 11.53
  & 10.56
  & 609.98
& 1,446.76
& 353.51
  & 7,486.65
& 757.75
& 7.10
& 8.57 \\

\textit{Ogbn-arxiv}
  & 0.02
  & 14.36
  & 14.05
  & 2,895.06
& 6,076.18
& 686.12
  & 2,687.45
& 1,685.18
& 9.96
& 12.45 \\

\textit{Ogbn-products}
  & 0.02
  & 517.95
  & 513.36
  & OOM
& OOM
& OOM
  & 71,489.00
& OOM
& 146.82
& 542.61 \\

\textit{Reddit}
  & 0.02
  & 24.40
  & 24.84
  & 2,672.85& 6,130.46& 337.15
  & 6,610.70& 1,815.77& 4.91& 11.50 \\

\bottomrule
\end{tabular}}
\end{table*}
\begin{table*}[ht]
\centering
\caption{Optimal hyperparameter settings for each dataset. The columns indicate whether incremental k-means++ ("If Incre.") and dual HPO ("If Dual HPO") were used (True/False), along with the best values for the fuzziness parameter, number of repetitions, the $\alpha$ coefficients, and weight decay ("wd").}
\resizebox{\linewidth}{!}{
\begin{tabular}{lcccccccc}
\toprule
\textbf{Dataset} & If Incre. & If Dual HPO & Fuzziness & \# Rep. & $\alpha_1$ & $\alpha_2$ & $\alpha_0$ & wd \\ 
\midrule
\textit{Citeseer}      & True  & True  & 1.1/1.3   & 50 & 0.5/0.1   & 0.9 & -0.3/-0.1 & 1e-3 \\
\textit{Cora}          & True  & True  & 1.3       & 50 & 0.9/-0.3  & 0.9 & -0.1/0    & 5e-4 \\
\textit{Pubmed}        & True  & False & 1.1       & 50 & 0.3       & 0.9 & -0.2      & 5e-4 \\
\textit{Flickr}        & False & False & 1         & 1  & 0         & 0.6 & 0.3       & 1e-3 \\ 
\midrule
\textit{Ogbn-arxiv (-real)}    & False & False & 1         & 1  & 0.5       & 0   & 0.6       & 1e-3 \\
\textit{Ogbn-products} & True  & False & 1         & 1  & 0.1       & 0   & 0.9       & 1e-3 \\
\textit{Reddit}        & True  & False & 1         & 1  & 0.2       & 0.9 & 0.7       & 5e-4 \\
\bottomrule
\end{tabular}}
\end{table*}\label{tab:hyper}

\section{Additional Results}
\subsection{Performance and Efficiency}
For simplicity, Table~\ref{tab:main} omits the standard error and running time of coreset selection methods. we provide the full results here.

Figure~\ref{fig:accuracy_vs_time} presents the accuracy vs. time trade-off for Reddit. For the remaining three large datasets, we provide the corresponding results in Figure~\ref{fig:accuracy_vs_time_large_vertical}. 
The results align with our main findings, further confirming that GECC surpasses the baselines in both efficiency and scalability. It consistently maintains stable performance while effectively managing computational resources throughout graph evolution. Notably, on the large-scale Ogbn-products dataset, which contains over one million nodes, most GC methods fail, whereas GECC remains robust and continues to operate successfully.

\begin{figure*}[htbp]
  \centering
  \begin{subfigure}[b]{0.85\linewidth}
    \centering
    \includegraphics[width=0.8\linewidth]{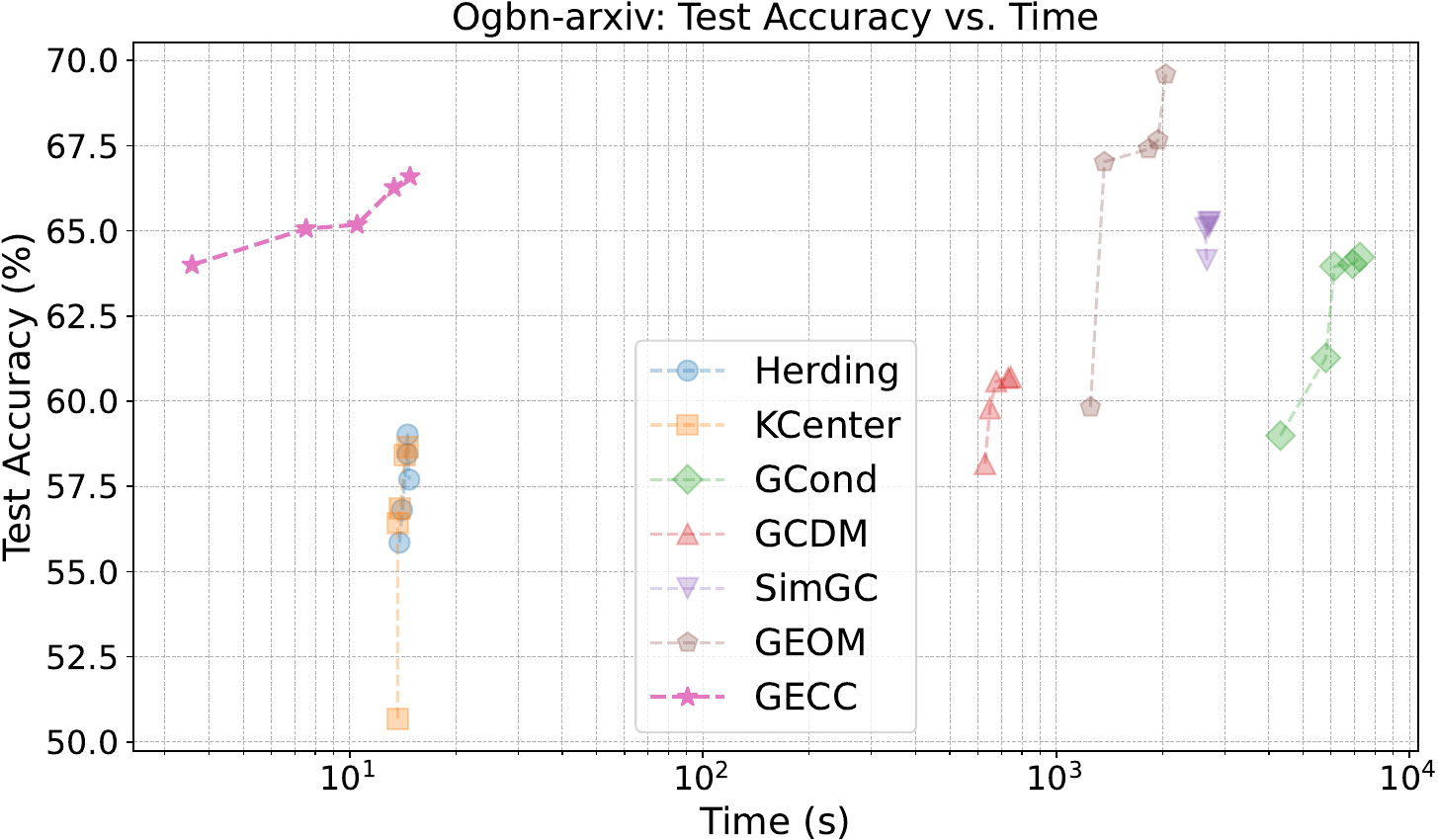}
    \label{fig:flickr}
  \end{subfigure}
  \begin{subfigure}[b]{0.85\linewidth}
    \centering
    \includegraphics[width=0.8\linewidth]{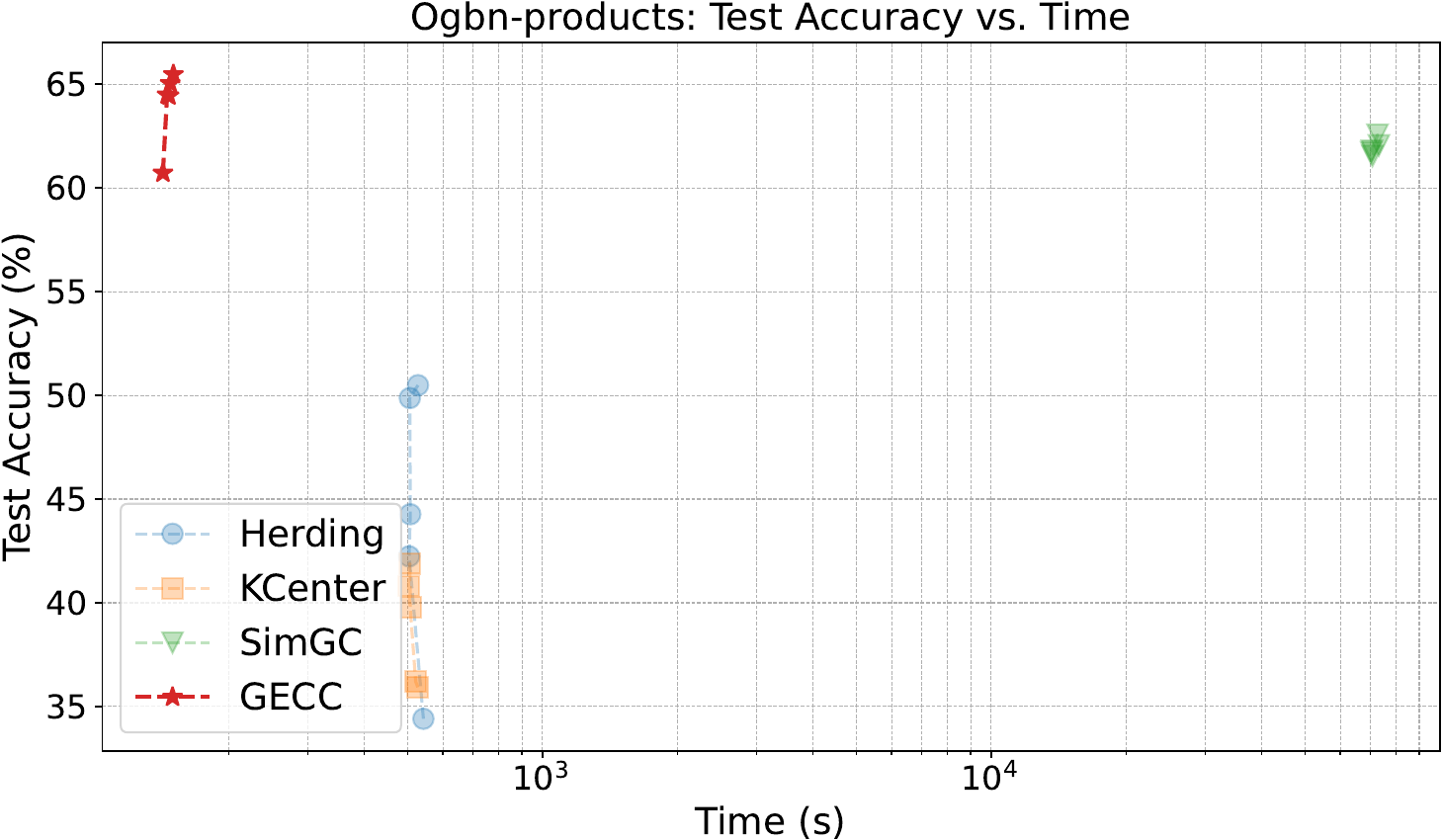}
    \label{fig:ogbnproducts}
  \end{subfigure}
  \begin{subfigure}[b]{0.85\linewidth}
    \centering
    \includegraphics[width=0.8\linewidth]{figs/Reddit_time_vs_accuracy-cropped.pdf}
    \label{fig:redproducts}
  \end{subfigure}
  \caption{Test accuracy vs. condensation time on large datasets (top-left is better).}
  \label{fig:accuracy_vs_time_large_vertical}
\end{figure*}

\subsection{Why We Choose the Depth=2 ?}
We pick $K=2$ based on findings from most deeper GNN papers. Below is a snippet for Cora, Ogbn-arxiv, and Reddit, showing that $K=2$ performs best. We add the following sensitivity analysis.

\begin{table}[ht]
\centering
\caption{Sensitivity Analysis of GECC Performance by Feature Propagation Depth. The best-performing depth for each dataset is highlighted in bold.}
\label{tab:depth_sensitivity}
\begin{tabular}{@{}lcccccc@{}}
\toprule
\textbf{Dataset} & \multicolumn{6}{c}{\textbf{Feature Propagation Depth}} \\
\cmidrule(l){2-7}
& \textbf{0} & \textbf{1} & \textbf{2} & \textbf{3} & \textbf{4} & \textbf{5} \\
\midrule
Cora       & 19.24 & 78.13 & \textbf{82.99} & 82.15 & 81.80 & 81.41 \\
Ogbn-arxiv & 63.64 & 66.60 & \textbf{66.71} & 66.13 & 66.24 & 66.29 \\
Reddit     & 90.44 & 90.44 & \textbf{91.37} & 91.35 & 91.28 & 91.14 \\
\bottomrule
\end{tabular}
\end{table}
\subsection{Visualization of Traceability}
Our t-SNE-based experiment color-codes original nodes by class and links them to their centroid(s), revealing:
\begin{itemize}
    \item \textbf{Clear class-wise alignment}: Synthetic nodes (cluster centroids) align well with the clusters of original nodes, indicating that GECC preserves the semantic structure of each class during condensation.
    \item \textbf{Compact cluster coverage}: Each synthetic node is well-embedded within the local neighborhood of its corresponding class, highlighting that the condensed representations accurately cover the original data distribution.
    \item \textbf{Few synthetic outliers}: Unlike many condensation approaches that generate scattered or poorly positioned centroids, our method yields minimal outlier centroids, suggesting robust and stable clustering even in high-dimensional spaces.
    \item \textbf{Scalability and consistency}: Even in large-scale datasets, our synthetic nodes remain representative and well-distributed, showing the scalability of our clustering method without sacrificing membership fidelity.
\end{itemize}
\begin{figure*}[t]
    \centering
    \includegraphics[width=\linewidth]{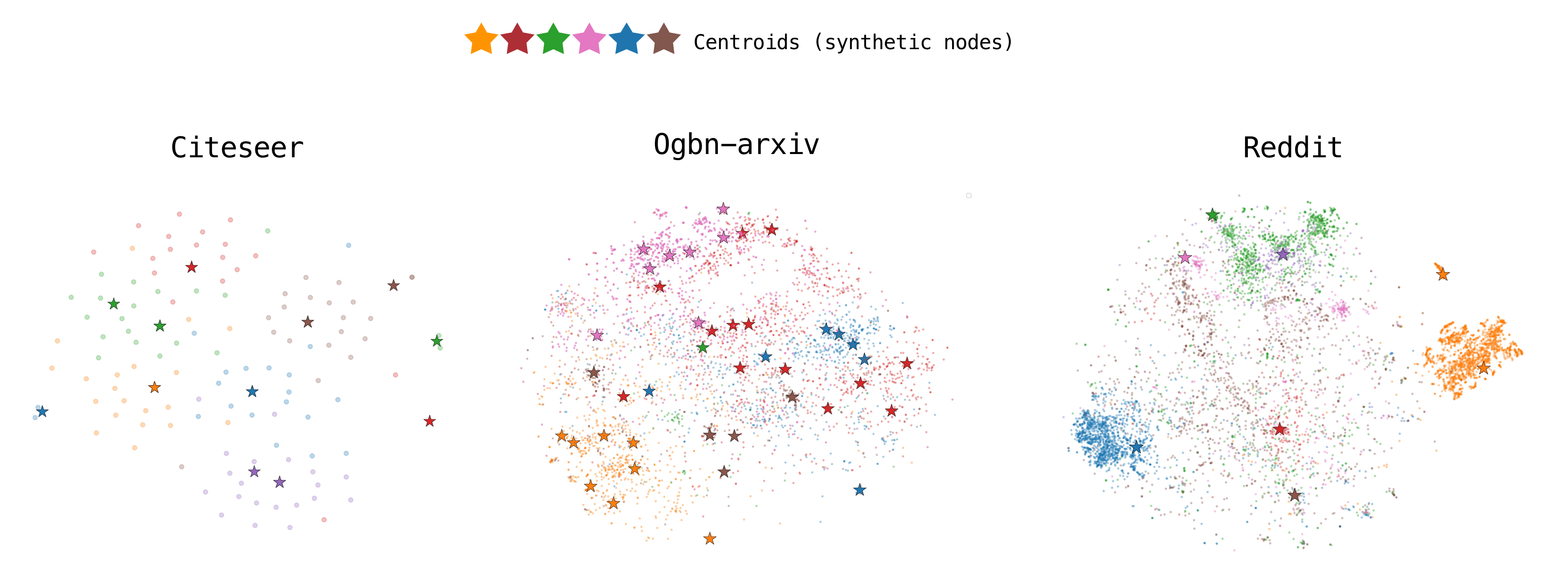}
    \caption{T-SNE results of the GECC synthetic node embeddings and original node embeddings across different classes.}
    \label{fig:placeholder}
\end{figure*}

\section{Discussions}
\subsection{How the Theory Extend to Non-linear GNNs?}
While our theoretical analysis focuses on SGC, we believe local approximations could generalize our approach to non-linear models. However, non-linearities introduce complexities like non-convex optimization and intricate layer interactions, making closed-form guarantees difficult. Many prior works~\cite{gao2024rethinking,jin2022condensing} similarly rely on linear or quasi-linear assumptions. We aim to broaden our framework to fully accommodate non-linear GNNs in future work.
\subsection{Clarification of Our Evolving Setting Compared to Incremental Learning} Both our work and incremental learning work such as OpenGC~\cite{opengc} address evolving datasets, yet the settings differ in important ways. Specifically, our paper focuses on batch-wise evolution, where \textbf{the training set is expanded or updated in batches from the same data source.} By contrast, OpenGC deals with a task-wise evolution scenario and is centered on out-of-distribution issues.

\subsection{When to Use Hard or Soft Clustering in GECC?}
Hard clustering is faster (each node has a single cluster) and suits large datasets or real-time scenarios. Soft clustering allows fractional memberships but is more costly. As more efficient soft clustering emerges, it may become viable for large graphs and potentially improve GECC.

\end{document}